\documentclass[english]{article}

\usepackage{proceed2e}
\usepackage[margin=1in]{geometry}
\usepackage{times}
\usepackage{subcaption}

\RequirePackage{fancyhdr}
\RequirePackage{color}
\RequirePackage{algorithm}
\RequirePackage{algorithmic}

\RequirePackage{natbib}

\usepackage[resetlabels,labeled]{multibib}
\newcites{sup}{Supplementary References}

\RequirePackage{forloop}

\usepackage{paralist}
\usepackage{ifxetex}

\ifxetex
\usepackage{fontspec}
\else
\usepackage[T1]{fontenc}
\usepackage[utf8]{inputenc}
\fi

\usepackage{amsmath}
\usepackage{amsfonts}
\usepackage{amssymb}

\newcommand{\norm}[1]{\left\lVert#1\right\rVert}

\usepackage{booktabs}
\usepackage{array}
\usepackage{multirow}

\usepackage{lipsum}
\usepackage{adjustbox}

\usepackage{appendix}
\usepackage{hyperref}

\usepackage{bm}
\usepackage{tikz}
\usetikzlibrary{calc,trees,positioning,arrows,chains,shapes.geometric,%
  decorations.pathreplacing,decorations.pathmorphing,shapes,%
  matrix,shapes.symbols,fit,decorations,arrows.meta,%
  fit,shapes.misc}

\usepackage{xcolor}
\definecolor{nice-red}{HTML}{E41A1C}
\colorlet{dark-red}{nice-red!80!black}
\definecolor{nice-orange}{HTML}{FF7F00}
\colorlet{dark-orange}{orange!85!black}
\definecolor{nice-yellow}{HTML}{FFC020}
\definecolor{nice-green}{HTML}{4DAF4A}
\definecolor{nice-blue}{HTML}{377EB8}
\definecolor{nice-purple}{HTML}{984EA3}

\newcolumntype{L}[1]{>{\raggedright\let\newline\\\arraybackslash\hspace{0pt}}m{#1}}
\newcolumntype{C}[1]{>{\centering\let\newline\\\arraybackslash\hspace{0pt}}m{#1}}
\newcolumntype{R}[1]{>{\raggedleft\let\newline\\\arraybackslash\hspace{0pt}}m{#1}}

\newcommand{\Real}{\ensuremath{\mathbb{R}}}

\newcommand{\Complex}{\ensuremath{\mathbb{C}}}

\newcommand{\fs}{\ensuremath{\phi}}
\newcommand{\z}{\ensuremath{\mathbf{z}}}

\DeclareMathOperator{\proj}{proj}

\DeclareMathOperator{\findAdversarialSet}{FindAdversarialSet}
\DeclareMathOperator{\trainDiscriminator}{TrainDiscriminator}
\DeclareMathOperator{\mainLoop}{AdversarialSetTraining}

\newcommand{\x}{\ensuremath{\mathbf{x}}}
\newcommand{\params}{\ensuremath{\boldsymbol{\theta}}}
\newcommand{\hparams}{\ensuremath{\boldsymbol{\gamma}}}
\newcommand{\h}{\ensuremath{\mathbf{h}}}

\newcommand{\AdvSet}{\ensuremath{\mathcal{S}}}
\newcommand{\Ass}{\ensuremath{\mathcal{A}}}

\newcommand{\subspace}{\ensuremath{\mathcal{U}}}

\newcommand{\kg}[1]{{\textsc{#1}}}
\newcommand{\card}[1]{{|#1|}}

\newcommand{\rel}[1]{{\textsc{#1}}}

\newcommand{\ie}{\textit{i}.\textit{e}.}
\newcommand{\eg}{\textit{e}.\textit{g}.}

\newcommand{\ents}{\ensuremath{\mathcal{E}}}
\newcommand{\rels}{\ensuremath{\mathcal{R}}}
\newcommand{\vars}{\ensuremath{\mathcal{V}}}

\newcommand{\Implies}{\ensuremath{\Rightarrow}}

\newcommand{\res}[1]{\textsc{#1}}

\newcommand{\mdl}[1]{{\textsc{#1}}}

\newcommand{\tdot}[3]{\ensuremath{\langle #1, #2, #3 \rangle}}

\newcommand{\ReP}[1]{\ensuremath{\text{Re}\left(#1\right)}}
\newcommand{\ImP}[1]{\ensuremath{\text{Im}\left(#1\right)}}
\newcommand{\conjt}[1]{\ensuremath{\overline{#1}}}

\newcommand{\loss}{\ensuremath{\mathcal{J}}}

\newcommand{\factLoss}{\ensuremath{\mathcal{J}_{\mathcal{F}}}}
\newcommand{\advLoss}{\ensuremath{\mathcal{J}_{\mathcal{I}}}}
\newcommand{\advLossMax}{\ensuremath{\mathcal{J}^{\max}_{\mathcal{I}}}}
\newcommand{\lossMax}{\ensuremath{\mathcal{J}^{\max}}}

\newcommand{\KG}{\ensuremath{\mathcal{G}}}
\newcommand{\corr}[1]{\ensuremath{\delta(#1)}}

\usepackage{xspace}
\newcommand{\ASR}{ASR\xspace}

\newcommand{\mASR}{\emph{ASR}}
\newcommand{\mcASR}{\emph{cASR}}

\usepackage{url}

\usepackage[colorinlistoftodos]{todonotes}

\usepackage{menukeys}
\usepackage{comment}

\makeatother

\begin{document}

\title{Adversarial Sets for Regularising Neural Link Predictors}

\author{
    Pasquale Minervini$^1$ \, Thomas Demeester$^2$ \, Tim Rocktäschel$^3$ \, Sebastian Riedel$^1$ \\
University College London, London, United Kingdom$^1$ \\
Ghent University - iMinds, Ghent, Belgium$^2$ \\
University of Oxford, Oxford, United Kingdom$^3$\\
}

\maketitle

\begin{abstract}
In adversarial training, a set of models learn together by pursuing competing goals, usually defined on \emph{single} data instances.
However, in relational learning and other non-\emph{i.i.d} domains, goals can also be defined over \emph{sets} of instances.
For example, a link predictor for the $\rel{is-a}$ relation needs to be consistent with the \emph{transitivity} property: if $\rel{is-a}(x_{1}, x_{2})$ and $\rel{is-a}(x_{2}, x_{3})$ hold, $\rel{is-a}(x_{1}, x_{3})$ needs to hold as well.
Here we use such assumptions for deriving an \emph{inconsistency loss}, measuring the degree to which the model violates the assumptions on an adversarially-generated set of examples.
The training objective is defined as a minimax problem, where an \emph{adversary} finds the most offending adversarial examples by maximising the inconsistency loss, and the model is trained by jointly minimising a supervised loss and the inconsistency loss on the adversarial examples.
This yields the first method that can use function-free Horn clauses (as in Datalog) to regularise any neural link predictor, with complexity independent of the  domain size.
We show that for several link prediction models, the optimisation problem faced by the adversary has efficient closed-form solutions.
Experiments on link prediction benchmarks indicate that given suitable prior knowledge, our method can significantly improve neural link predictors on all relevant metrics.
\end{abstract}
\section{INTRODUCTION}

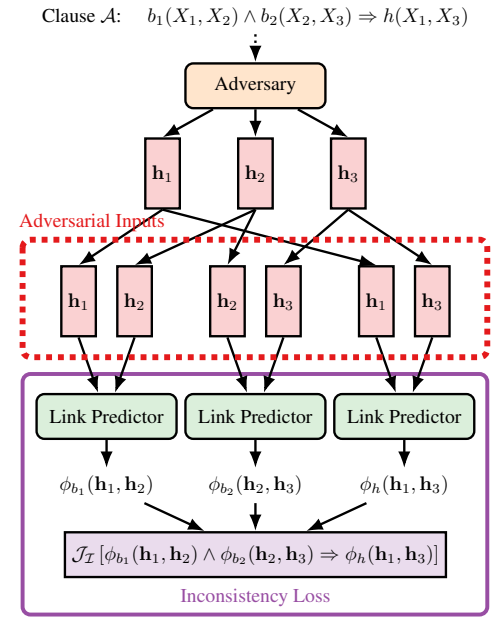
\begin{figure}[t!]
    \centering
    \begin{adjustbox}{width=.8\columnwidth}
    \begin{tikzpicture}[x=1cm,y=1cm, very thick, auto]
        \tikzstyle{embedding} = [draw, fill=nice-blue!20, minimum width=.6cm, minimum height=1.25cm]
        \tikzstyle{virtual} = [embedding, fill=nice-red!20]
        \tikzstyle{adversary} = [draw, fill=nice-orange!20, rounded corners, minimum width=2.5cm, minimum height=.8cm]
        \tikzstyle{predictor} = [draw, fill=nice-green!20, rounded corners, minimum width=2.5cm, minimum height=.8cm]
        \tikzstyle{loss} = [draw, fill=nice-purple!20, minimum width=2.5cm, minimum height=.8cm]

        \draw[line width=2pt, nice-purple, rounded corners] (-1,-1.3) rectangle (7.3,-5.6);

        \node[virtual] (e11) {$\h_1$};
        \node[virtual, right of = e11] (e12) {$\h_2$};

        \node[virtual, right = 1cm of e12] (e22) {$\h_2$};
        \node[virtual, right of = e22] (e23) {$\h_3$};

        \node[virtual, right = 1cm of e23] (e31) {$\h_1$};
        \node[virtual, right of = e31] (e33) {$\h_3$};

        \node[] (pe22) at ($(e22)!0.5!(e23)$) {};
        \node[virtual, above = 1.5cm of pe22] (e2) {$\h_2$};
        \node[virtual, left = 1cm of e2] (e1) {$\h_1$};
        \node[virtual, right = 1cm of e2] (e3) {$\h_3$};

        \node[adversary, above = .5cm of e2] (adversary) {Adversary};

        \node[above = 0.5cm of adversary] (clause) {Clause $\Ass$: \quad $b_1(X_1, X_2) \land b_2(X_2, X_3) \Implies h(X_1, X_3)$};

        \node[] (lp1p) at ($(e11)!0.5!(e12)$) {};
        \node[] (lp2p) at ($(e22)!0.5!(e23)$) {};
        \node[] (lp3p) at ($(e31)!0.5!(e33)$) {};

        \node[predictor, below = 1.5cm of lp1p] (lp1) {Link Predictor};
        \node[predictor, below = 1.5cm of lp2p] (lp2) {Link Predictor};
        \node[predictor, below = 1.5cm of lp3p] (lp3) {Link Predictor};

        \node[below = .5cm of lp1] (s1) {$\fs_{b_1}(\h_1, \h_2)$};
        \node[below = .5cm of lp2] (s2) {$\fs_{b_2}(\h_2, \h_3)$};
        \node[below = .5cm of lp3] (s3) {$\fs_{h}(\h_1, \h_3)$};

        \node[loss, below = .5cm of s2] (loss) {$\advLoss\left[\fs_{b_1}(\h_1, \h_2) \land \fs_{b_2}(\h_2, \h_3) \Implies \fs_{h}(\h_1, \h_3)\right]$};

        \node[below = 0.05cm of loss, nice-purple] {Inconsistency Loss};

        \draw[-Latex, dotted] (clause) -- (adversary);
        \draw[-Latex] (adversary) -- (e1.north);
        \draw[-Latex] (adversary) -- (e2.north);
        \draw[-Latex] (adversary) -- (e3.north);

        \draw[-Latex] (e1.south) -- (e11.north);
        \draw[-Latex] (e1.south) -- (e31.north);

        \draw[-Latex] (e2.south) -- (e12.north);
        \draw[-Latex] (e2.south) -- (e22.north);

        \draw[-Latex] (e3.south) -- (e23.north);
        \draw[-Latex] (e3.south) -- (e33.north);

        \draw[-Latex] (e11.south) -- (lp1);
        \draw[-Latex] (e12.south) -- (lp1);

        \draw[-Latex] (e22.south) -- (lp2);
        \draw[-Latex] (e23.south) -- (lp2);

        \draw[-Latex] (e31.south) -- (lp3);
        \draw[-Latex] (e33.south) -- (lp3);

        \draw[-Latex] (lp1) -- (s1);
        \draw[-Latex] (lp2) -- (s2);
        \draw[-Latex] (lp3) -- (s3);

        \draw[-Latex] (s1) -- (loss);
        \draw[-Latex] (s2) -- (loss);
        \draw[-Latex] (s3) -- (loss);

        \draw[line width=3pt, nice-red, rounded corners, dashed] (-1,1.1) rectangle (7.3,-1);
        \node[anchor = west, nice-red] at (-1.2,1.4) {Adversarial Inputs};
    \end{tikzpicture}
    \end{adjustbox}
    \caption{Overview of regularisation via adversarial sets for neural link prediction. Given a clause $\Ass$ with variables $\{ X_1, X_2, X_3 \}$, an adversary maps each variable to an adversarial entity embedding that maximises an inconsistency loss, while the discriminator is trained by minimising the link prediction and inconsistency loss.}
    \label{fig:architecture}
\end{figure}
Adversarial training~\citep{DBLP:conf/kdd/DalviDMSV04,DBLP:conf/nips/GoodfellowPMXWOCB14,szegedy14:intriguing} is a learning setting where two or more models learn together by pursuing competing goals.
Adversarial learning has received increasing attention in recent years, mainly motivated by the observation that neural networks can be vulnerable to carefully-crafted adversarial examples~\citep{DBLP:conf/nips/GoodfellowPMXWOCB14}.
Adding additional loss terms to training objectives for enforcing models to be robust to adversarial examples can be very beneficial, for instance in the context of image synthesis~\citep{dosovitskiy2015learning}.
Such competing goals are generally defined on \emph{single} data instances.
For example, in Generative Adversarial Networks (GANs)~\citep{DBLP:conf/nips/GoodfellowPMXWOCB14}, a generator is trained to synthesise single \emph{fake} data instances (\eg{} images) that are classified as \emph{real} by a discriminator, while the discriminator is trained to discriminate between single \emph{real} and \emph{fake} data instances.
However, for relational tasks such as \emph{link prediction} or \emph{knowledge base population}~\citep{nickel2012factorizing,riedel2013relation,socher2013reasoning,chang2014typed,toutanova2015representing,neelakantan2015compositional}, where objects can interact with each other, such goals can also be defined in terms of several instances.
For instance, consider the hypernymy relation $\rel{is-a}$.
Since state-of-the-art link predictors rank or score pairs in isolation~\citep{DBLP:conf/nips/BordesUGWY13,toutanova2015representing,DBLP:conf/icml/TrouillonWRGB16}, they cannot explicitly account for the transitivity property of $\rel{is-a}$.
This means that a predictor might infer $\rel{is-a}(x_{1}, x_{2})$ and $\rel{is-a}(x_{2}, x_{3})$, but not  $\rel{is-a}(x_{1}, x_{3})$.
In this case, encouraging transitivity to hold for such models can be achieved by defining a goal on three related inputs rather than a single input: $\{ (x_{1}, x_{2}), (x_{2}, x_{3}), (x_{1}, x_{3})\}$.
The goal of the adversary would be to find inputs that lead to inconsistent predictions, while the predictor's goal would be restoring consistency on such sets of inputs.
In this paper, we introduce Adversarial Set Regularisation (\ASR), a general and scalable method for regularising neural link prediction models by using background knowledge.
Our method is summarised by the \emph{adversarial training} architecture in Fig.~\ref{fig:architecture}.
In \ASR, we first define a set of constraints on multiple problem instances in the form of function-free First-order Logic (FOL) clauses. From these clauses we then derive an \emph{inconsistency loss} that measures the extent to which constraints are violated.
The learning architecture is composed of two models, an \emph{adversary} and a \emph{discriminator}, having two competing goals.
The \emph{adversary} tries to find a set of adversarial input representations for which, according to the discriminator (a link prediction model), the constraints do not hold. Such a set is found by maximising the inconsistency loss.
The \emph{discriminator}, on the other hand, uses the inconsistency loss on the adversarial input representations for regularising its training process.
Our proposed training algorithm, described in Sect.~\ref{sec:asets}, can be seen as a zero-sum game involving two players, where:
\begin{inparaenum}[i)]
\item one player, the link predictor, has to predict a target graph over a set of \emph{real} inputs, while ensuring global consistency over a set of \emph{generated adversarial} inputs; and
\item the other player, the adversary, has to generate adversarial input representations such that the link predictor fails at constructing consistent graphs.
\end{inparaenum}
The link prediction model is trained by jointly minimising both the data loss and the inconsistency loss on the adversarial input sets by updating the model, while the adversary maximises the inconsistency loss by changing the input sets.
Another interpretation is that the adversary acts as an \emph{adaptive regulariser} for training neural link predictors.
Our method is related to \citet{DBLP:conf/naacl/RocktaschelSR15} and its variant \mdl{KALE}~\citep{DBLP:conf/emnlp/GuoWWWG16}, which also minimise an inconsistency loss over sets of instances.
A core difference with the proposed approach is that, instead of generating adversarial examples in representation space, they select random entities from the training and test sets.
This generally leads to less effective updates, since the random entities can already satisfy the imposed constraints.
Furthermore, they also enforce consistency not just by changing the relation-specific parameters, but also by changing the distributed representations of the entity inputs. This can lead to poor generalisation on entities less frequently sampled during training.
\citet{DBLP:conf/emnlp/DemeesterRR16} overcome this problem, but their approach only works for a single type of model and simple clauses.
Our approach has no such restrictions, and we formally show that it is a generalisation of their work.
We show empirically that, by training on adversarial input sets, the link prediction model becomes indeed more robust.
In experiments, we use our approach to inject prior assumptions into several state-of-the-art models, indicating the general nature of our approach.
The regularised models outperform the original models on real-world data sets, namely WN18 and FB122.
Moreover, using the same constraints and experimental setup, we outperform \mdl{KALE} both in terms of absolute performance, and in terms of relative improvements when compared to their underlying base models.
Our contributions are threefold:
\begin{inparaenum}[i)]
\item we introduce a novel approach to regularise neural link prediction models based on prior relational assumptions, such as transitivity -- this is the first work that uses adversarial input sets for doing so;
\item we present an optimisation algorithm for solving the underlying minimax problem; and
\item we derive closed form solutions for the inner maximisation problem that enable faster training, provide intuitive insights into the goal of the adversary, and show that the method of \citet{DBLP:conf/emnlp/DemeesterRR16} can be derived from our framework.
\end{inparaenum}
In the next sections we first briefly introduce the link prediction problem and state-of-the-art link prediction models.
Then we present our adversarial approach to injecting prior knowledge. We conclude with experimental results and a discussion.
\section{LINK PREDICTION}
In this work, we focus on the problem of \emph{predicting missing links} in large, multi-relational networks such as \kg{Freebase}.
In the literature, this problem is referred to as \emph{link prediction}.
We specifically focus on \emph{knowledge graphs}, \ie{}, graph-structured knowledge bases where factual information is stored in the form of relationships between entities.
Link prediction in knowledge graphs is also known as \emph{knowledge base population}.
We refer to \citet{DBLP:journals/pieee/Nickel0TG16} for a recent survey on approaches for this problem.
A knowledge graph $\KG \triangleq \{ (r, a_{1}, a_{2}) \} \subseteq \rels \times \ents \times \ents$ can be formalised as a set of triples (facts) consisting of a relation type $r \in \rels$ and two entities $a_1, a_2 \in \ents$, respectively referred to as the \emph{subject} and the \emph{object} of the triple.
Each triple $(r, a_{1}, a_{2})$ encodes a relationship of type $r$ between $a_1$ and $a_2$, represented by the fact $r(a_1, a_2)$.
Link prediction in knowledge graphs is often simplified to a \emph{learning to rank} problem, where the objective is to find a score or ranking function $\fs^{\params}_r : \ents \times \ents \mapsto\Real$ for a relation $r$ that can be used for ranking triples according to the likelihood that the corresponding facts hold true.
\subsection{Neural Link Prediction}
Recently, a specific class of link predictors received a growing interest~\citep{DBLP:journals/pieee/Nickel0TG16}. These predictors can be understood as multi-layer neural networks.
Given a triple $\x = (r, a_1, a_2)$, the associated score $\fs^{\params}_r(a_1,a_2)$ is given by a neural network architecture encompassing an \emph{encoding layer} and a \emph{scoring layer}.
In the encoding layer, the subject and object entities $a_1$ and $a_2$ are mapped to low-dimensional vector representations (embeddings) $\h_1 \triangleq \h(a_1) \in \Real^k$ and $\h_2 \triangleq \h(a_2) \in \Real^k$, produced by an encoder $\h^{\hparams} : \ents \to \Real^k$ with parameters $\hparams$.
This layer can be pre-trained~\citep{DBLP:conf/acl/VylomovaRCB16} or, 
more commonly, learnt from data by back-propagating the link prediction error to the encoding layer~\citep{DBLP:conf/nips/BordesUGWY13,DBLP:journals/pieee/Nickel0TG16,DBLP:conf/icml/TrouillonWRGB16}.
In the scoring layer, the entity representations $\h_1$ and $\h_2$ are scored by a function $\phi^{\params}_r(\h_1, \h_2)$, parametrised by $\params$.
Summarising, the high-level architecture is defined as:
\begin{equation*}
 \begin{aligned}
  \h_1, \h_2 && \triangleq & \quad \h^{\hparams}(a_1), \h^{\hparams}(a_2) \\
  \fs_r(a_{1}, a_{2}) && \triangleq & \quad \fs^{\params}_{r}(\h_1, \h_2)
 \end{aligned}
\end{equation*}
Ideally, more likely triples should be associated with higher scores, while less likely triples should be associated with lower scores.
While the literature has produced a multitude of encoding and scoring strategies, for brevity we overview only a small subset of these. However, we point out that our method makes no further assumptions about the network architecture other than the existence of an argument encoding layer.

\subsection{Encoding Layer}
Given an entity $a \in \ents$, the entity encoder $\h^{\hparams}$ is usually implemented as a simple embedding layer $\h^{\hparams}(a) \triangleq [ \hparams ]_a$, where $\hparams$ is an embedding matrix~\citep{DBLP:journals/pieee/Nickel0TG16}.
For pre-trained embeddings, the embedding matrix is fixed.
Note that other encoding mechanisms are conceivable, such as recurrent or convolutional neural networks.

\subsection{Scoring Layer}

\paragraph{DistMult}
\mdl{DistMult}~\citep{yang15:embedding} represents each relation $r$ using a parameter vector $\params_r \in \Real^{k}$, and scores a link of type $r$ between $(\h_1, \h_2)$ using the following scoring function:
\begin{equation*}
\fs_r^{\params}(\h_1,\h_2) \triangleq \tdot {\params_r}{\h_1}{\h_2} \triangleq \sum_{i = 1}^{k} \params_{r, i} \h_{1, i} \h_{2, i},
\end{equation*}
\noindent where $\tdot {\cdot}{\cdot}{\cdot}$ denotes the tri-linear dot product.

\paragraph{ComplEx}

\mdl{ComplEx}~\citep{DBLP:conf/icml/TrouillonWRGB16} is an extension of \mdl{DistMult} using complex-valued embeddings while retaining the mathematical definition of the dot product.
In this model, the scoring function is defined as follows:
\begin{equation*}
 \begin{aligned}
 \fs_r^{\params}(\h_1,\h_2) & \triangleq \ReP{\tdot{\params_r}{\h_1}{\conjt{\h_2}}},
 \end{aligned}
\end{equation*}
\noindent where $\params_r, \h_1, \h_2 \in \Complex^{k}$ are complex vectors, $\conjt{\mathbf{x}}$ denotes the complex conjugate of $\mathbf{x}$, and $\ReP{\mathbf{x}} \in \Real^{k}$  denotes the real part of $\mathbf{x}$.
\subsection{Training}
Training neural link predictors amounts to minimising a loss function defined over a target graph $\KG$ of triples $\{(r, a_1, a_2)\}$.
Since such graphs usually only contain positive examples (true facts), a common strategy is to generate negative examples by \emph{corrupting} the triples in the graph~\citep{rendle2009bpr,DBLP:conf/nips/BordesUGWY13,yang15:embedding,DBLP:journals/pieee/Nickel0TG16}.
Formally, given a triple $(r, a_1, a_2)$, negative examples are generated by a corruption process $\delta$ defined by:
\begin{equation*}
 \begin{aligned}
  &\corr{r, a_1, a_2} \triangleq\nonumber\\
  &\qquad\{ (r, \tilde{a_1}, a_2) \mid \tilde{a_1} \in \ents \} \cup \{ (r, a_1, \tilde{a_2}) \mid \tilde{a_2} \in \ents \}.
 \end{aligned}
\end{equation*}
The main motivation for this negative sampling strategy is the Local Closed World Assumption (LCWA)~\citep{DBLP:conf/kdd/0001GHHLMSSZ14}.
According to the LCWA, if a triple exists in the graph, other triples obtained by corrupting either the subject or the object of the triples not appearing in the graph can be considered as negative examples.
Similarly to~\citet{DBLP:conf/nips/BordesUGWY13}, we learn the model parameters $\params$ and $\hparams$ by minimising a \emph{hinge loss} \factLoss{}, referred to as \emph{fact loss}, defined over positive and negative examples:
\begin{align}
&\factLoss(\KG ; \params, \hparams) \triangleq \label{eq:factloss}\\
&\sum_{(r, a_1, a_2) \in \Omega} \left[ 1 - y_{r}(a_{1}, a_{2}) \cdot \fs^{\params}_r(\h^{\hparams}(a_1), \h^{\hparams}(a_2)) \right]_{+}, \nonumber
\end{align}
\noindent where $\mathcal{N} \triangleq \{ \tilde{\x} \mid \x \in \KG \land \tilde{\x} \in \corr{\x} \}$ is the set of negative examples (triples) generated by corrupting the triples in $\KG$, $\Omega \triangleq \KG \cup \mathcal{N}$ is a set containing both positive and negative examples, and $y_{r}(a_1, a_2) = \pm 1$ is an indicator function with value $1$ if $(r, a_1, a_2) \in \KG$, and $-1$ otherwise.
For several neural link prediction models, the fact scores can be trivially increased by increasing the magnitude of entity embeddings.
A common solution is to either regularise the entity representations~\citep{yang15:embedding,DBLP:conf/icml/TrouillonWRGB16}, or require them to live in subspaces such as the unit cube~\citep{DBLP:conf/emnlp/DemeesterRR16} $\{ \h \mid \h \in \left[0, 1\right]^{k} \}$, or in the unit sphere~\citep{DBLP:conf/nips/BordesUGWY13} $\{ \h \mid \norm{\h}^{2}_{2} = 1 \}$.
\section{ADVERSARIAL SETS} \label{sec:asets}
Even though the training data may be consistent with various assumptions we can make about the graph, on sets of unseen triples the local nature of the classifiers may still lead to inconsistencies.
Taking the \rel{is-a} example, we may see a high score for $(\rel{is-a}, \res{cat}, \res{feline})$ and $(\rel{is-a}, \res{feline}, \res{animal})$, but a low score for $(\rel{is-a}, \res{cat}, \res{animal})$, violating the transitivity property of the $\res{is-a}$ hypernymy relation.
To address this problem, we generate \emph{adversarial input sets}, and encourage the model to fix its inconsistencies with respect to these inputs.
More specifically, we find a set of adversarial entity embeddings as inputs to the scoring layer of the model, instead of a set of actual entity pairs.
This has two core benefits.
First, it allows us to solve a continuous optimisation problem (over embeddings) as opposed to a combinatorial one (over actual entities). The former can even have closed form solutions, as shown in Sect.~\ref{ssec:closed}.
Second, it forces the model to learn general correlations between relations, as opposed to knowledge about specific facts and entities through the encoder.
For clarity, we now consider a single assumption $\Ass$, such as the transitivity of relation $r$. Generalising to multiple assumptions only requires instantiating one adversary and one inconsistency loss for each of the assumptions.
In this work, we use Horn Clauses, a subset of FOL formulae, to express our assumptions.
For example, transitivity of the hypernym relation can be expressed by:
\begin{equation} \label{eq:clause}
\rel{is-a}(X_1, X_2) \land \rel{is-a}(X_2, X_3) \Implies \rel{is-a}(X_1, X_3),
\end{equation}
\noindent where the atom on the right-hand side of the implication is referred to as the \emph{head} of the clause, the conjunction of atoms on the left-hand side is referred to as the \emph{body} of the clause, and all variables are universally quantified.
Let an \emph{adversarial input set} $\AdvSet$ define a mapping from the free variables in $\vars$ in $\Ass$ to $k$-dimensional embeddings -- \ie{} $\AdvSet : \vars \mapsto \Real^{k}$.
We call $\AdvSet$ a \emph{set} because it implicitly specifies a set of adversarial inputs to the scoring layers of the neural link predictors associated with the atoms in $\Ass$.
For example, in the case of the transitivity clause in Eq. (\ref{eq:clause}), a mapping $\AdvSet$ with $\AdvSet(X_1) = \h_1$, $\AdvSet(X_2) = \h_2$ and $\AdvSet(X_3) = \h_3$ will define the set of inputs $(\h_1, \h_2)$, $(\h_2, \h_3)$ and $(\h_1, \h_3)$ to the scoring layer $\fs_{\rel{is-a}}$.
Given the adversarial input set $\AdvSet$, the \emph{inconsistency loss} $\advLoss(\Ass ; \params, \AdvSet)$ measures the degree to which assumption $\Ass$ is violated on $\AdvSet$ with model parameters $\params$. It is computed as a function of the neural link prediction scores on the adversarial inputs in $\AdvSet$.
For the transitivity clause in Eq. (\ref{eq:clause}), the inconsistency loss is computed as a function of $\fs_{\rel{is-a}}(\h_1, \h_2)$, $\fs_{\rel{is-a}}(\h_2, \h_3)$ and $\fs_{\rel{is-a}}(\h_1, \h_3)$.
Our loss function is then a linear combination of the fact loss function in Eq. (\ref{eq:factloss}) and the inconsistency loss $\advLoss$:
\begin{equation*}
\loss(\KG, \Ass; \params, \hparams, \AdvSet) \triangleq \factLoss(\KG; \params, \hparams) + \alpha \advLoss(\Ass ; \params, \AdvSet),
\end{equation*}
\noindent where $\alpha \in \Real$ controls the extent to which the assumption $\Ass$ should be enforced.
Our adversarial training algorithm attempts to find input sets $\AdvSet$ with maximal inconsistency, and model parameters $\params, \hparams$ that minimise such an inconsistency.
This can be formalised by the following minimax problem:
\begin{equation}\label{adv-opt}
\min_{\params, \hparams} \max_{\AdvSet} \loss(\KG, \Ass; \params, \hparams, \AdvSet).
\end{equation}
Note that the search over possible $\AdvSet$ needs to take into account the unit sphere or cube constraints mentioned earlier: for any variable $X_i \in \vars$, the corresponding $k$-dimensional embedding $\AdvSet(X_i)$ should live on the same subspace (\eg{} unit sphere or unit cube) as the entity embeddings.
To instantiate this framework we need to be able to map an assumption $\Ass$ to an inconsistency loss $\advLoss(\Ass ; \params, \AdvSet)$, and to solve the optimisation problem in Eq. (\ref{adv-opt}).
\subsection{Inconsistency Losses}
Given an assumption $\Ass$ expressed as a FOL clause $\res{body} \Implies \res{head}$, as in Eq. (\ref{eq:clause}), our goal is defining a loss term $\advLoss(\AdvSet)$ that assesses the degree to which $\Ass$ is violated on a set of adversarial inputs $\AdvSet$.
Recall that we represent $\AdvSet : \vars \mapsto \Real^{k}$ as a binding of the free variables $\vars$ in $\Ass$ to adversarially-trained embeddings in $\Real^{k}$.
This means that in practice we have to search over variable-to-embedding bindings.
We construct the inconsistency loss $\advLoss$ compositionally, by first calculating $\fs(\res{head})$ and $\fs(\res{body})$, respectively representing the scores for the head and body of the clause, based on the binding of variables to embeddings defined by $\AdvSet$.
Subsequently, we test whether the head score is lower than the body score, \ie{}, $\fs(\res{head}) < \fs(\res{body})$~\citep{DBLP:conf/emnlp/DemeesterRR16}.
If so, we assign a penalty proportional to the margin between body score and the head score.
This yields the following inconsistency loss:
\begin{equation*}
 \advLoss(\res{body} \Implies \res{head}) \triangleq \left[ \fs(\res{body}) - \fs(\res{head}) \right]_{+}.
\end{equation*}
The motivation for this loss is that implications can be understood as “whenever the body is true, the head has to be true as well”.
In terms of neural link prediction models, this translates into “the score of the head should at least be as large as the score of the body”.
For calculating the inconsistency loss, we need to specify how to calculate the scores of the head and body.
To score a single atom, we simply map the free variables with the corresponding embeddings contained in the adversarial set $\AdvSet$ and apply the neural link predictor scoring function:
\begin{equation*}
 \fs \left( r\left(X_i, X_j\right) \right) = \fs_{r} \left(\AdvSet(X_i), \AdvSet(X_j)\right). 
\end{equation*}
This gives us the score of the head atom, and the scores of the atoms within the body.
Similarly to the product t-norm used in \citet{DBLP:conf/naacl/RocktaschelSR15}, we use the G{\"o}del t-norm, a continuous generalisation of the conjunction operator in logic~\citep{Gupta:1991:TTN:107687.107690} to score the body of a clause, \ie{}, a conjunction of several atoms:
\begin{equation*}
\fs(A \land B) \triangleq \min\{ \fs(A), \fs(B) \},
\end{equation*}
\noindent where $A$ and $B$ are clause atoms.
This allows us to back-propagate through a conjunction of atoms.
For disjunction one can use $\fs(A \lor B) \triangleq \max\{ \fs(A), \fs(B) \}$, and for negation $\fs(\neg A) \triangleq - \fs(A)$, which allows the use of arbitrary function-free FOL clauses as in~\cite{DBLP:conf/naacl/RocktaschelSR15}.
However, in our experiments we only use Horn clauses, and leave the investigation of more complex clauses for future work.
\subsection{Optimisation}
\begin{algorithm}[t]
	\caption{Solving the minimax problem in Eq. (\ref{adv-opt}) via Projected Stochastic Gradient Descent} \label{alg:sgd}
	\begin{algorithmic}[1]
	\small
	    \REQUIRE{No. of epochs $\{ \tau_{a}, \tau_{d}, \tau \}$, learning rates $\{ \pmb{\eta}^{a}, \pmb{\eta} \}$}
		\MAIN{$\mainLoop(\Ass)$}
		\STATE{Randomly initialise $\{ \params_{0}, \hparams_{0} \}$}
		\FOR{$i \in \{ 1, \ldots, \tau \}$}
		    \STATE{$\AdvSet_{i} \leftarrow \findAdversarialSet(\Ass, \params_{i - 1})$}
		    \STATE{$(\params_{i}, \hparams_{i}) \leftarrow \trainDiscriminator(\KG, \params_{i - 1}, \hparams_{i - 1}, \AdvSet_{i})$}
		\ENDFOR
		\STATE{\textbf{return} $\params_{\tau}, \hparams_{\tau}$}
		\ENDMAIN
		\FUNCTION{$\findAdversarialSet(\Ass, \params)$}
		\STATE{Randomly initialise $\AdvSet_{0}$}
		\FOR{$i = 1, \ldots, \tau_{a}$}
    		\STATE{$\h_j \leftarrow \proj(\h_j), \; \forall \h_j \in \AdvSet$}
    		\STATE{$g_{i} \leftarrow  \nabla_{\AdvSet} \advLoss(\Ass ; \params, \AdvSet_{i - 1})$}
    		\STATE{$\AdvSet_{i} \leftarrow \AdvSet_{i - 1} + \eta^{a}_{i} g_{i}$}
		\ENDFOR
		\STATE{\textbf{return} $\AdvSet_{\tau_{a}}$}
		\ENDFUNCTION
		\FUNCTION{$\trainDiscriminator(\KG, \params, \hparams, \AdvSet)$}
	    \STATE{$\params_{0} \leftarrow \params, \hparams_{0} \leftarrow \hparams$}
		\FOR{$i = 1, \ldots, \tau_{d}$}
    		\STATE{$\h_{j} \leftarrow \proj(\h_{j}), \; \forall j \in \{ 1, \ldots, \card{\ents} \}$}
    		\STATE{$g_{i} \leftarrow  \nabla_{\langle \params, \hparams \rangle} \loss(\KG, \Ass ; \params_{i - 1}, \hparams_{i - 1}, \AdvSet)$}
    		\STATE{$(\params_{i}, \hparams_{i}) \leftarrow (\params_{i - 1}, \hparams_{i - 1}) - \eta_{i} g_{i}$}
		\ENDFOR
		\STATE{\textbf{return} $\params_{\tau_{d}}, \hparams_{\tau_{d}}$}
		\ENDFUNCTION
	\end{algorithmic}
\end{algorithm}
To optimise the minimax objective in Eq. (\ref{adv-opt}), we alternate between two optimisation processes, as shown in Alg. \ref{alg:sgd}.
On line 4, the algorithm finds an adversarial set $\AdvSet$ by maximising the inconsistency loss using $\tau_{a}$-many Gradient Ascent iterations.
On line 5, the link prediction model is trained by jointly minimising the fact loss and the inconsistency loss on the adversarial input set $\AdvSet$ via $\tau_{d}$-many Stochastic Gradient Descent iterations.
In our implementation of Alg. \ref{alg:sgd}, we enforce all entity and adversarial embeddings to live either on the unit cube, \ie{}, $\proj(\h) \triangleq \min(\max(\h, \mathbf{0}), \mathbf{1})$, or on the unit sphere, \ie{}, $\proj(\h) \triangleq \h/\norm{\h}_{2}$.
At the beginning of the training process, we initialise the neural link predictor parameters $\{ \params, \hparams \}$ using uniform Xavier initialisation~\citep{DBLP:journals/jmlr/GlorotB10}.
When searching for the adversarial set $\AdvSet$ that maximises the inconsistency loss, we initialise $\AdvSet$ using randomly sampled entity embeddings.
Note that this algorithm is independent of the specific neural link prediction model used, and applicable to any function-free FOL clause.

\subsection{Closed Form Solutions} \label{ssec:closed}
\begin{figure}[t]
\caption{$\advLossMax$ contours for simple implications with \mdl{ComplEx}, for varying $\params_{b, i} - \params_{r, i}$, for unit sphere (left) and unit cube (right) constraints.}
\begin{subfigure}{.23\textwidth}
  \centering {
   \includegraphics[width=1\textwidth]{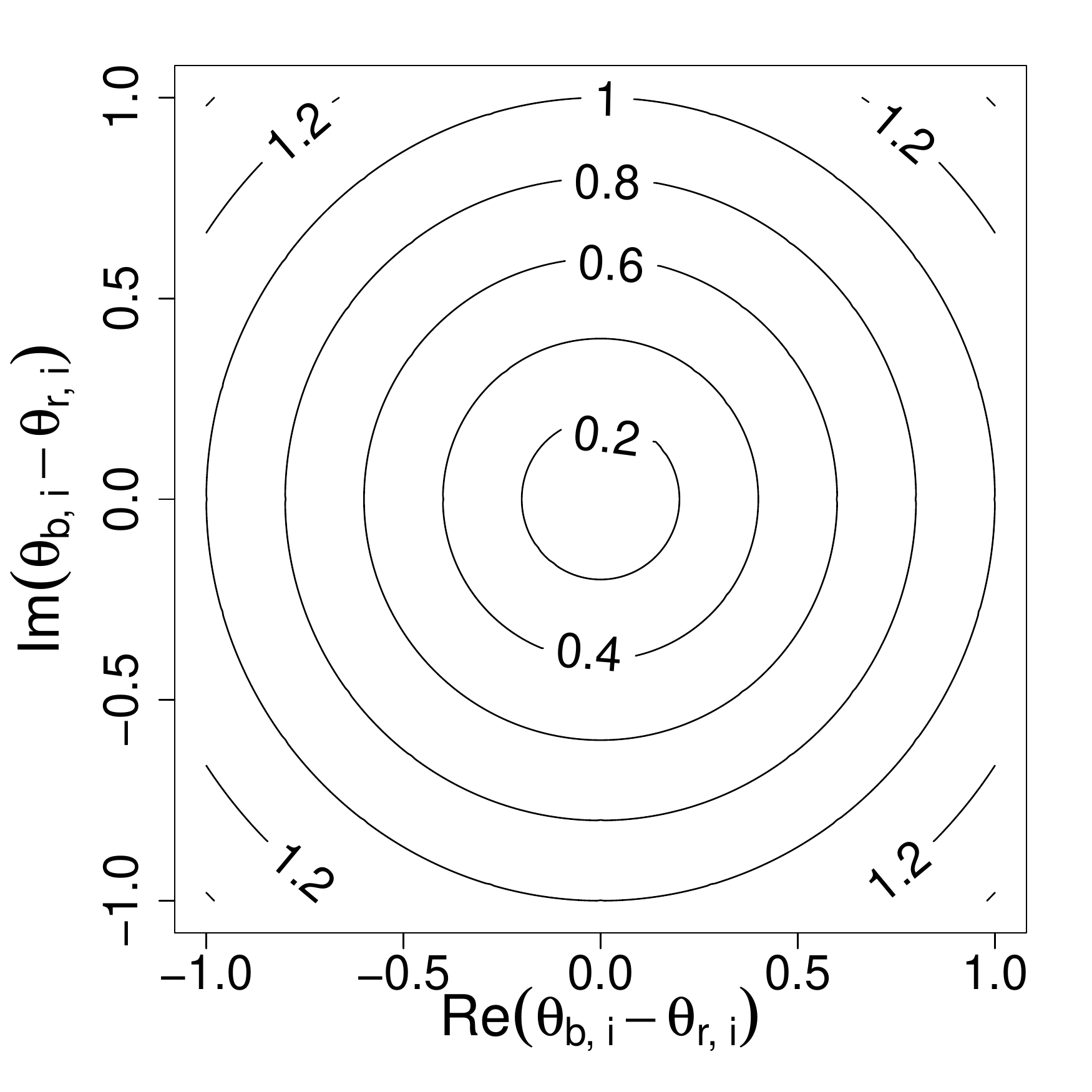}
   \caption{Unit Sphere} \label{fig:sfig1}
  }
\end{subfigure}
\hfill
\begin{subfigure}{.23\textwidth}
  \centering{
   \includegraphics[width=1\textwidth]{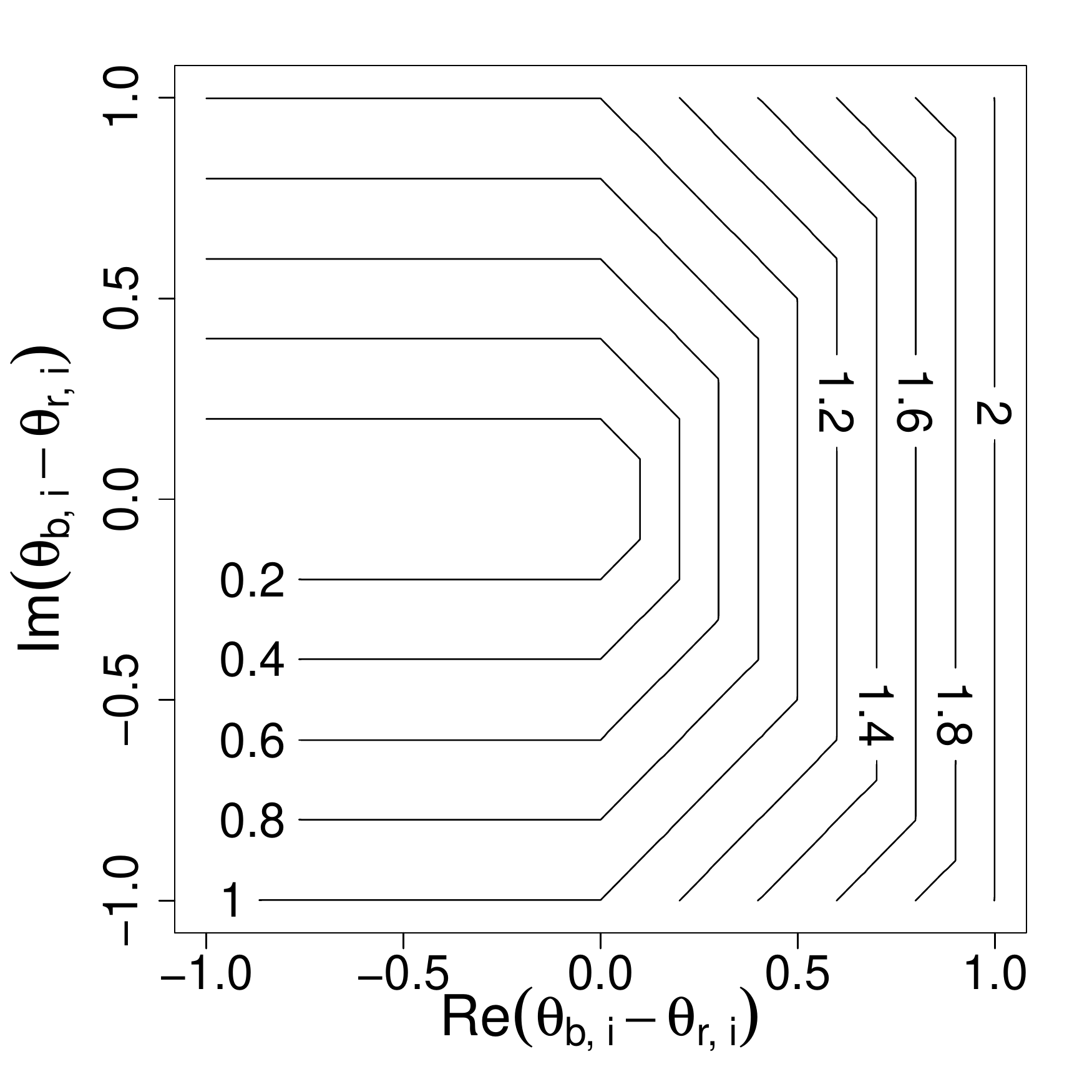}
   \caption{Unit Cube} \label{fig:sfig2}
  }
\end{subfigure}
\label{fig:contours}
\vspace{-12pt}
\end{figure}

\begin{table}[t!]
\centering
\caption{
Closed form expressions $\advLossMax$ for \mdl{DistMult} and \mdl{ComplEx} on three types of clauses.
For \mdl{DistMult} $\boldsymbol{\delta} = \params_{b} - \params_{r} \in\mathbb{R}^k$, and for \mdl{ComplEx}, $\boldsymbol{\delta} = \params_{b} - \params_{r} = \overline{\boldsymbol{\zeta}} \in\mathbb{C}^k$, for entity embeddings restricted to the unit sphere (left) and unit cube (right) subspaces.} \label{tab:closedformsolutions}
\vspace{1em}
\begin{adjustbox}{width=1\columnwidth}
\begin{tabular}{ccccc}
\toprule
{\bf Clause} & {\bf Model} & {\bf Unit Sphere} & {\bf Unit Cube} \\
\midrule
\multirow{ 2}{*}{ $\begin{array} {l@{}} r(X_1, X_2) \\  \quad\Implies r(X_2, X_1) \end{array}$ } & \mdl{DistMult}
& $0$
& $0$ \\
& \mdl{ComplEx}
& $\max_i \left\{2\vert \params_{r, i}^{\text{I}}\vert \right\}$
& $2\sum_i \vert \params_{r, i}^{\text{I}}\vert$
\\
\midrule
\multirow{ 2}{*}{ $\begin{array} {l@{}} b(X_1, X_2) \\  \quad\Implies r(X_1, X_2) \end{array}$ } & \mdl{DistMult}
& $\max_i \left\{ \vert\delta_i\vert \right\}$
& $\sum_{i} \max\left\{0, \delta_i\right\}$ \\
& \mdl{ComplEx}
& $\max_i \left\{\sqrt{(\delta_i^{\text{R}})^2 + (\delta_i^{\text{I}})^2 } \right\}$
&
$\sum_i \max(0,{\delta}_i^{\text{R}}) + \max({\delta}_i^{\text{R}},\vert{\delta}_i^{\text{I}}\vert)$
\\
\midrule
\multirow{ 2}{*}{ $\begin{array} {l@{}} b(X_1, X_2) \\  \quad\Implies r(X_2, X_1) \end{array}$ } & \mdl{DistMult}
& $\max_i \left\{ \vert\delta_i\vert \right\}$
& $\sum_{i} \max\left\{0, \delta_i\right\}$ \\
& \mdl{ComplEx}
& $\max_j \left\{\sqrt{(\zeta_i^{\text{R}})^2 + (\zeta_i^{\text{I}})^2 } \right\}$
&
$\sum_i \max(0,{\zeta}_i^{\text{R}}) + \max({\zeta}_i^{\text{R}},\vert{\zeta}_i^{\text{I}}\vert)$
\\
\bottomrule
\end{tabular}
\end{adjustbox}
\end{table}
While the algorithm above is much more efficient than grounding out the clauses over the entity space (as in \cite{DBLP:conf/naacl/RocktaschelSR15}), it still requires the inner loop of maximising the inconsistency loss.
Compared to closed-form analytical solutions, the inner optimisation loop can be computationally less efficient, require more hyperparameters (learning rate, number of iterations etc.) and offer fewer guarantees on whether the global optimum is found.
In some cases, it is possible to analytically calculate the solution $\advLossMax(\Ass ; \params)$ to the inner optimisation problem of maximising the inconsistency loss $\advLoss(\Ass ; \params, \AdvSet)$ with respect to the adversarial set $\AdvSet$:
\begin{equation*}
 \advLossMax(\Ass ; \params) \triangleq \max_\AdvSet  \advLoss(\Ass ; \params, \AdvSet).
\end{equation*}
When $\advLossMax$ is known up front, the inner training loop disappears. We will call this approach the \emph{Closed-Form} Adversarial Set Regularisation (\emph{cASR}) method, as opposed to the more general iterative method in Alg.~\ref{alg:sgd}.
We derive $\advLossMax$ for several types of clauses, both for \mdl{DistMult} and \mdl{ComplEx}, as shown in Tab.~\ref{tab:closedformsolutions}.
Full derivations are provided in the supplementary material.
Let's consider the closed-form inconsistency loss $\advLossMax$ for simple clauses of the form $b(X_1, X_2) \Implies r(X_1, X_2)$, and let $\params_{b}, \params_{r}$ denote the predicate embeddings of predicates $b$ and $r$, respectively.
From Tab.~\ref{tab:closedformsolutions} we can see that the expressions in the unit sphere case are independent of the sign of $\params_{b, i} - \params_{r, i}$, indicating that the non-symmetric implications cannot be explicitly modelled with unit sphere constraints.
Both our theoretical and experimental results indicate that unit cube constraints used by \citet{DBLP:conf/emnlp/DemeesterRR16} are indeed better suited for rule injection.
Also note that the lifted clause injection method by \citet{DBLP:conf/emnlp/DemeesterRR16} can be seen as a special case of this formulation, which is however limited to a single neural link prediction model and a small subset of clauses.
Fig.~\ref{fig:contours} shows contours of the contribution to $\advLossMax$ from component $\params_{b, i} - \params_{r, i}$, for simple implications with \mdl{ComplEx}.
We observe the rotation-symmetric behaviour for the unit sphere constraints (Fig.~\ref{fig:sfig1}). In contrast, for unit cube constraints (Fig.~\ref{fig:sfig2}) we see the quite different impact of the real part (on its own a scaled version of the expression for \mdl{DistMult}) and that the imaginary part contributes as soon as its absolute value exceeds the real part.
Minimising $\advLossMax$ for unit sphere constraints in \mdl{ComplEx} boils down to encouraging $\params_{r, i} \approx \params_{b, i}$, for both real and imaginary parts, whereas for unit cube constraints it encourages an ordering relation of the real parts $\ReP{\params_{r, i}} \geq \ReP{\params_{b, i}}$, and the equality of the imaginary parts $\ImP{\params_{r, i}} \approx \ImP{\params_{b, i}}$.
%


%
\section{RELATED WORK} 
\citet{rocktaschel14low,DBLP:conf/naacl/RocktaschelSR15} provide a framework for jointly maximising the probability of observed facts and propositionalised First-Order Logic clauses.
\citet{DBLP:conf/ijcai/WangWG15} show how different types of clauses can be included after training the model by using Integer Linear Programming.
Recently, \citet{DBLP:conf/ijcai/WangC16} propose a method for embedding facts and clauses using matrix factorisation.
However, all these approaches ground the clauses in the training data. This limits their scalability to comparably small knowledge bases that contain only few entities.
As mentioned in~\citet{DBLP:conf/emnlp/DemeesterRR16}, such problems provide an important motivation for \emph{lifted} clause injection methods that do not rely on grounding of clauses, but rather regularise relation representations directly and ensure that the assumptions hold on the whole entity embedding space, improving generalisation.
\citet{DBLP:conf/kdd/0001GHHLMSSZ14,DBLP:conf/nips/NickelJT14} and \citet{DBLP:conf/ijcai/WangWG15} propose combining observable patterns in the form of clauses and latent features for link prediction tasks.
However, in these models, clauses are not used for learning better distributed representations of entities and relation types.
Our model is related to \mdl{Model FSL} proposed by~\citet{DBLP:conf/emnlp/DemeesterRR16}, where they use very simple clauses for defining a partial ordering among relation embeddings.
Moreover, their approach is limited to the simple matrix factorisation \mdl{Model F} by \citet{DBLP:conf/naacl/RiedelYMM13}, while \ASR extends and improves over \mdl{Model FSL}.
First, it can be used for injecting arbitrarily complex first-order logic clauses, such as the transitivity clause in Eq. (\ref{eq:clause}).
Second, it can be used jointly with any knowledge graph embedding model.
Furthermore, it also improves over the general clause-injection methods by \citet{DBLP:conf/naacl/RocktaschelSR15} and \citet{DBLP:conf/emnlp/GuoWWWG16} since it avoids generating possible groundings of a clause.
Gaifman models~\citep{DBLP:conf/nips/Niepert16} and related works train latent representations using features derived from subregions of a knowledge base: in such models, one can define relationships between subregions.
%


%
\section{EXPERIMENTS} \label{sec:experiments}
\begin{table*}
\caption{Examples of the clauses used for \kg{Freebase} (FB122) and \kg{WordNet} (WN18).} \label{tab:rules}
\centering
%
%
\begin{adjustbox}{width=1\textwidth}
\begin{tabular}{lr}
\toprule
\multicolumn{2}{l}{$\rel{/people/person/nationality}(X_1,X_2) \land \rel{/location/country/official\_language}(X_2,X_3) \Implies \rel{/people/person/languages}(X_1,X_3)$} \\
\multicolumn{2}{l}{$\rel{/country/administrative\_divisions}(X_1,X_2) \land \rel{/administrative\_division/capital}(X_2,X_3) \Implies \rel{/location/contains}(X_1,X_3)$} \\
\multicolumn{2}{l}{$\rel{/location/country/capital}(X_1,X_2) \Implies \rel{/location/location/contains}(X_1,X_2)$} \\
\midrule
$\rel{\_hypernym}(X_1, X_2) \Implies \rel{\_hyponym}(X_2, X_1)$ & $\rel{\_hyponym}(X_1, X_2) \Implies \rel{\_hypernym}(X_2, X_1)$ \\
$\rel{\_has\_part}(X_1, X_2) \Implies \rel{\_part\_of}(X_2, X_1)$ & $\rel{\_part\_of}(X_1, X_2) \Implies \rel{\_has\_part}(X_2, X_1)$ \\
\bottomrule
\end{tabular}
\end{adjustbox}
\vspace{-8pt}
\end{table*}
\begin{table}[t]
	\caption{Statistics for the data sets.} \label{tab:datasets}
	\centering {
		\begin{adjustbox}{width=1\columnwidth}
			\begin{tabular}{rcccccc}
				\toprule
				{\bf Dataset} & $\card{\ents}$ & $\card{\rels}$ & \#{\bf Train} & \#{\bf Valid} & \#{\bf Test-I} & \#{\bf Test-II} \\
				\midrule
				\kg{\bf FB122} & 9,738 & 122 & 91,638 & 9,595 & 5,057 & 6,186 \\
				\kg{\bf WN18} & 40,943 & 18 & 141,442 & 5,000 & 1,394 & 3,606 \\
				\bottomrule
			\end{tabular}
        \end{adjustbox}
	}
	\vspace{-8pt}
\end{table}
We now present our experimental results, starting by describing the experimental setup and hyper-parameters, both on synthetic and real-world data-sets.
All models were implemented in TensorFlow, and source code is available on-line.\footnote{\url{https://github.com/uclmr/inferbeddings}}
\paragraph{Experimental Setup for Synthetic Data}
In order to investigate the effect of \ASR on particular types of clauses, we created the following small synthetic knowledge base.
We randomly sampled subject-object pairs from a low-dimensional entity space (with probability 0.1, and for $\card{\ents} = 30$).
With each of these pairs, we created facts by combining them with 15 relations in $\rels$ with probability 0.1, resulting in 135 training facts.
Besides those, we created clauses by randomly combining relations into the appropriate clause template, \ie{} symmetry constraints, implications, and transitivity.
For each experiment on a particular type of clause, 10 different clauses of that type were applied together.
The training facts contain no evidence of these clauses.
Each new fact that could be inferred based on the clauses was added as a positive item to the test data, together with a negative example obtained by corrupting its subject-object pair.
\paragraph{Experimental Setup for Real Data}
We also evaluate the proposed method on \kg{WordNet} (WN18) and \kg{Freebase} (FB122) jointly with the set of rules released by \citet{DBLP:conf/emnlp/GuoWWWG16}.
\kg{WordNet}~\citep{DBLP:journals/cacm/Miller95} is a lexical knowledge base for the English language, where entities correspond to word senses, and relationships define lexical relations between them.
The WN18 dataset consists of a subset of \kg{WordNet}, containing 40,943 entities, 18 relation types, and 151,442 triples.
\kg{Freebase}~\citep{DBLP:conf/aaai/BollackerCT07} is a large knowledge graph that stores general facts about the world.
The FB122 dataset is a subset of \kg{Freebase} regarding the topics of \emph{people}, \emph{location} and \emph{sports}, and contains 9,738 entities, 122 relation types, and 112,476 triples.
For both data sets, we used the fixed training, validation, test sets and rules provided by \citet{DBLP:conf/emnlp/GuoWWWG16}; a subset of the rules is shown in Tab.~\ref{tab:rules}.
Note that a subset of the test triples can be inferred by deductive logic inference.
For such a reason, following \citet{DBLP:conf/emnlp/GuoWWWG16}, we also partition the test set in two subsets, namely Test-I and Test-II: Test-I contains triples that \emph{cannot} be inferred by deductive logic inference, while Test-II contains all remaining test triples.
Statistics for the data sets are shown in Tab.~\ref{tab:datasets}.
\paragraph{Hyper-parameters}
We ran a grid search for the embedding dimension $k \in \{ 20, 50, 100, 150, 200 \}$, the margin $\gamma \in \{ 1, 2, 5, 10 \}$, the weight of the adversarial loss $\alpha \in \{ 1, 10, 10^{2}, 10^{3}, 10^{4} \}$, the number of iterations for the discriminator and the adversary $\tau_{d}, \tau_{a} \in \{ 1, 10 \}$, fixed the number of epochs to $\tau = 100$, and decided the optimal subspace $\subspace$ among either unit cube or unit sphere.
We use AdaGrad~\citep{DBLP:journals/jmlr/DuchiHS11} for automatically selecting the optimal learning rate, with an initial value of $0.1$.
For the synthetic dataset experiments, we use the same settings, fixing the hyper-parameters to $k=20$, $\gamma=1$, $\alpha=1$, $\tau_d= 10$ and $\tau_a=1$.
\paragraph{Evaluation Metrics}
For evaluating each model, we measure the quality of the ranking of test triples in terms of \emph{Mean Reciprocal Rank} (MRR) and \emph{HITS@$k$}~\citep{DBLP:conf/nips/BordesUGWY13,DBLP:journals/pieee/Nickel0TG16}.
MRR and Hits@$K$ are two standard evaluation measures on these data sets and are used in two settings: \emph{raw} and \emph{filtered}~\citep{DBLP:conf/nips/BordesUGWY13}.
Our results are reported in the \emph{filtered} setting, where metrics are computed after removing all the other known triples appearing in the training, validation or test sets from the ranking.
This is motivated by observing that ranking a test triple lower than another true triple should not be penalised.
For the experiments on synthetic data, we calculate the Area Under the Precision-Recall Curve (AUC-PR), based on the test data consisting of true facts and a single corruption of each true fact.
Each experiment is repeated 10 times with different randomly generated train facts, test facts, and clauses.
The reported AUC-PR values are averaged over these runs.
%

%
%
\section{RESULTS AND DISCUSSION} 
%
%
%
%
\begin{table*}
\centering
\caption{Link prediction results on the Test-I, Test-II and and Test-ALL on FB122, filtered setting.} \label{tab:testres}
\begin{adjustbox}{width=1\textwidth}
\begin{tabular}{C{1cm}lC{1cm}C{1cm}C{1cm}C{1cm}C{.5cm}C{1cm}C{1cm}C{1cm}C{1cm}C{.5cm}C{1cm}C{1cm}C{1cm}C{1cm}}
\toprule
& & \multicolumn{4}{c}{Test-I} & & \multicolumn{4}{c}{Test-II} & & \multicolumn{4}{c}{Test-ALL} \\
\cline{3-6}\cline{8-11}\cline{13-16}
& & \multicolumn{3}{c}{Hits@N (\%)} & \multirow{2}{*}{MRR} & & \multicolumn{3}{c}{Hits@N (\%)} & \multirow{2}{*}{MRR} & & \multicolumn{3}{c}{Hits@N (\%)} & \multirow{2}{*}{MRR} \\
\cline{3-5}\cline{8-10}\cline{13-15}
& & 3 & 5 & 10 & & \ & 3 & 5 & 10 & \ & & 3 & 5 & 10 & \ \\
\cline{2-16}
\multirow{9}{*}{\rotatebox[origin=c]{90}{\parbox[c]{1cm}{\centering {\bf FB122}}}}
& \mdl{TransE}~\citep{DBLP:conf/nips/BordesUGWY13} & 36.0 & 41.5 & 48.1 & 0.296 & & 77.5 & 82.8 & 88.4 & 0.630 & & 58.9 & 64.2 & 70.2 & 0.480 \\
%
& \mdl{KALE-Pre}~\citep{DBLP:conf/emnlp/GuoWWWG16} & 35.8 & 41.9 & 49.8 & 0.291 & & 82.9 & 86.1 & 89.9 & 0.713 & & 61.7 & 66.2 & 71.8 & 0.523 \\
& \mdl{KALE-Joint}~\citep{DBLP:conf/emnlp/GuoWWWG16} & {\bf 38.4} & {\bf 44.7} & {\bf 52.2} & 0.325 & & 79.7 & 84.1 & 89.6 & 0.684 & & 61.2 & 66.4 & 72.8 & 0.523 \\
\cline{2-16}

& \mdl{DistMult}~\citep{yang15:embedding} & 36.0 & 40.3 & 45.3 & 0.313 & & 92.3 & 93.8 & 94.7 & 0.874 & & 67.4 & 70.2 & 72.9 & 0.628 \\
& \emph{ASR}-\mdl{DistMult} & 36.3 & 40.3 & 44.9 & 0.330 & & 98.0 & 99.0 & 99.2 & 0.948 & & 70.7 & 73.1 & 75.2 & 0.675 \\
& \emph{cASR}-\mdl{DistMult} & 37.0 & 40.4 & 45.1 & 0.337 & & 96.7 & 98.6 & 99.3 & 0.933 & & 70.1 & 72.7 & 75.1 & 0.669 \\
\cline{2-16}

& \mdl{ComplEx}~\citep{DBLP:conf/icml/TrouillonWRGB16} & 37.0 & 41.3 & 46.2 & 0.329 & & 91.4 & 91.9 & 92.4 & 0.887 & & 67.3 & 69.5 & 71.9 & 0.641 \\
& \emph{ASR}-\mdl{ComplEx} & 37.3 & 41.0 & 45.9 & 0.338 & & {\bf 99.2} & {\bf 99.3} & {\bf 99.4} & {\bf 0.984} & & {\bf 71.7} & {\bf 73.6} & {\bf 75.7} & {\bf 0.698}
\\
& \emph{cASR}-\mdl{ComplEx} & 37.9 & 41.7 & 46.2 & {\bf 0.339} & & 97.7 & {\bf 99.3} & {\bf 99.4} & 0.954 & & 71.1 & {\bf 73.6} & 75.6 & 0.680 \\
\bottomrule
%
%
%
\end{tabular}
\end{adjustbox}
\vspace{-12pt}
\end{table*}
\begin{table}
\centering
\caption{Results on WN18 with limited training data -- \ie{} 20\%, 30\%, 40\% and 50\% of the training set.} \label{tab:limited}
\begin{adjustbox}{width=1\columnwidth}
\begin{tabular}{C{.5cm}clC{1cm}C{1cm}C{1cm}C{1cm}} 
\toprule
\multicolumn{3}{c}{\bf Training data} & 20\% & 30\% & 40\% & 50\% \\ 
\midrule
\multirow{4}{*}{\rotatebox[origin=c]{90}{\parbox[c]{1cm}{\centering {\bf WN18}}}}
\ & \multirow{2}{*}{Hits@10} & \mdl{ComplEx} & 38.9 & 46.6 & 54.1 & 60.5 \\ 
\ & \ & \mASR-\mdl{ComplEx} & {\bf 39.9} & {\bf 47.1} & {\bf 54.8} & {\bf 61.1} \\ 
\cline{2-7}
\ & \multirow{2}{*}{MRR} & \mdl{ComplEx} & 0.356 & 0.418 & 0.480 & 0.538 \\ 
\ & \ & \mASR-\mdl{ComplEx} & {\bf 0.366} & {\bf 0.423} & {\bf 0.484} & {\bf 0.540} \\ 
\bottomrule

\end{tabular}
\end{adjustbox}
\vspace{-8pt}
\end{table}
In this section, we describe our findings on the synthetic, \kg{WordNet} (WN18) and \kg{Freebase} (FB122) datasets.
\subsection{Synthetic Data} 
\paragraph{\ASR}
We tested the effectiveness of adversarial training for five different types of clauses in the synthetic data setup described in Section~\ref{sec:experiments}.
%
Table~\ref{tab:ITsynth} shows the resulting AUC-PR, where entity embeddings lie on the unit cube or on the unit sphere, and with vs.\ without iterative adversarial training ($\alpha = 0$ and $\alpha = 1$, respectively).
\mdl{DistMult} and \mdl{ComplEx} are able to encode the various types of clauses into the relation and argument representations. In line with arguments given above, the results for unit cube entities are generally better than their counterparts on the unit sphere, both on the standard models and with \ASR.
The most complex type of clauses (\ie{} the transitivity over three different relations) yields lower absolute AUC-PR values than simpler clauses.
Nonetheless, we observe a significant increase in scores due to the clauses.  
A noticeable case where the standard models cannot be improved is for \mdl{DistMult} on the clause that expresses symmetry (the first clause in Table~\ref{tab:closedformsolutions}).
This is unsurprising since symmetry is satisfied by construction in \mdl{DistMult}.
%
%
This experiment confirms that ASR is able to encode different types of clauses (unlike \cite{DBLP:conf/emnlp/DemeesterRR16}), and for different models (unlike \cite{DBLP:conf/naacl/RocktaschelSR15}).

\begin{table}[t!]
\centering
\caption{
AUC-PR results for \emph{ASR}-\mdl{DistMult} and \emph{ASR}-\mdl{ComplEx} on synthetic datasets with various types of clauses (with $r\not=s\not=t$). Comparison of standard models without clauses ($\alpha = 0$) and iterative adversarial training with clauses ($\alpha = 1$), with unit cube or unit sphere constraints on the entity embeddings. 
}
\label{tab:ITsynth}
\vspace{1em}
\begin{adjustbox}{width=1\columnwidth}
\begin{tabular}{llcccc}
\toprule
\multirow{ 2}{*}{\bf Clauses} & \multirow{ 2}{*}{\bf Model} & $\alpha = 0$ & $\alpha = 0$ & $\alpha = 1$ & $\alpha = 1$  \\
&& {\bf Unit Cube} & {\bf Unit Sphere} & {\bf Unit Cube} & {\bf Unit Sphere} \\
\midrule

\multirow{ 2}{*}{ $\begin{array} {l@{}} r(X_1, X_2) \\  \quad\Rightarrow r(X_2, X_1) \end{array}$ } & \mdl{DistMult} & \textbf{96.0} & 95.7 & 95.9 & 95.6 \\
 & \mdl{ComplEx} & 45.0 & 42.4 & 90.6 & \textbf{91.2} \\
\midrule
\multirow{ 2}{*}{ $\begin{array} {l@{}} r(X_1, X_2) \\  \quad\Rightarrow s(X_1, X_2) \end{array}$ } & \mdl{DistMult} & 59.2 & 61.3 & \textbf{86.3} & 79.2 \\
 & \mdl{ComplEx} & 60.8 & 61.2 & \textbf{85.8} & 79.7 \\
\midrule
\multirow{ 2}{*}{ $\begin{array} {l@{}} r(X_1, X_2) \\  \quad\Rightarrow s(X_2, X_1) \end{array}$ } & \mdl{DistMult} & 59.6 & 59.1 & \textbf{76.6} & 73.1 \\
 & \mdl{ComplEx} & 54.7 & 51.4 & \textbf{87.5} & 81.0 \\
\midrule
\multirow{ 2}{*}{$\begin{array} {l@{}} r(X_1, X_2) \wedge r(X_2, X_3) \\  \quad\Rightarrow r(X_1, X_3) \end{array}$} & \mdl{DistMult} & 45.0 & 37.5 & \textbf{65.9} & 45.8 \\
 & \mdl{ComplEx} & 40.9 & 36.3 & \textbf{54.8} & 42.5 \\
\midrule
\multirow{ 2}{*}{$\begin{array} {l@{}} r(X_1, X_2) \wedge s(X_2, X_3) \\  \quad\Rightarrow t(X_1, X_3) \end{array}$} & \mdl{DistMult} & 41.9 & 39.5 & 44.9 & \textbf{49.8} \\
 & \mdl{ComplEx} & 39.1 & 37.8 & 40.3 & \textbf{45.4} \\
\bottomrule
\end{tabular}
\end{adjustbox}
\end{table}

\paragraph{Closed-Form Solutions} 
Using the closed-form expressions for $\advLossMax$ given in Table~\ref{tab:closedformsolutions}, we performed the synthetic data set experiments again. The results, shown in Table~\ref{tab:CFsynth}, indicate that the optimal adversarial training is also able to encode clauses into the trained embeddings.
As with the iterative method, unit cube constraints perform better than on the unit sphere. For the synthetic data experiments, no hyper-parameter optimisation was performed, and we used the same number of discriminator training cycles as for the results presented in Table~\ref{tab:ITsynth}. Compared to the iterative method, closed form results on the unit sphere are consistently weaker. This is in line with our earlier observations that their symmetric character is not suited for modelling asymmetry in clauses. 
In any case, we observed a strong reduction in training time: the runs based on closed-form expressions took around a fifth the time of the corresponding iterative runs. 
\begin{table}[t!]
\centering
\caption{AUC-PR results on synthetic datasets for adversarial training with closed form expressions.}\label{tab:CFsynth}
\vspace{1em}
\begin{adjustbox}{width=1\columnwidth}
\begin{tabular}{L{3cm}lcc}
    \toprule
\multirow{2}{*}{\bf Clause} & \multirow{2}{*}{\bf Model} & $\alpha = 1$ & $\alpha = 1$  \\
\ & \ & {\bf Unit Cube} & {\bf Unit Sphere} \\
\midrule

\multirow{2}{*}{ $\begin{array} {l@{}} r(X_1, X_2) \\  \quad\Rightarrow r(X_2, X_1) \end{array}$ } & \mdl{DistMult} & \textbf{97.3} & 95.2 \\
 & \mdl{ComplEx} & \textbf{91.7} & 90.1 \\
\midrule
\multirow{2}{*}{ $\begin{array} {l@{}} r(X_1, X_2) \\  \quad\Rightarrow s(X_1, X_2) \end{array}$ } & \mdl{DistMult} & \textbf{85.0} & 69.5 \\
 & \mdl{ComplEx} & \textbf{83.6} & 75.5 \\
\midrule
\multirow{2}{*}{ $\begin{array} {l@{}} r(X_1, X_2) \\  \quad\Rightarrow s(X_2, X_1) \end{array}$ } & \mdl{DistMult} & \textbf{76.9} & 67.8 \\
 & \mdl{ComplEx} & \textbf{80.2} & 73.9 \\
\bottomrule
\end{tabular}
\end{adjustbox}
\vspace{-12pt}
\end{table}

\subsection{Link Prediction in Freebase and Wordnet}
We compare the proposed Adversarial Set Regularisation (\ASR) method with \mdl{KALE}, a recently proposed neural link prediction model that also leverages rules during the training process, the Translating Embeddings model (\mdl{TransE})~\citep{DBLP:conf/nips/BordesUGWY13}, \mdl{DistMult} and \mdl{ComplEx}.
In particular, we compare \ASR with two strong \mdl{KALE} variants: \mdl{KALE-Pre}, which augments training triples by means of logic inference, and \mdl{KALE-Joint}, which jointly considers training triples and rules during training~\citep{DBLP:conf/emnlp/GuoWWWG16}.
Results for \mdl{TransE} and \mdl{KALE} are reported from \citet{DBLP:conf/emnlp/GuoWWWG16}.
In experiments, we use \ASR for regularising the learning process in \mdl{DistMult} and \mdl{ComplEx} -- we denote the resulting models by \mASR-\mdl{DistMult} and \mASR-\mdl{ComplEx}: we retain the original formulations of the scoring functions $\fs_{r}$, but train the models by solving the minimax optimisation problem in Eq. (\ref{adv-opt}) using Alg.~\ref{alg:sgd}.

Results for \kg{Freebase} are reported in Tab.~\ref{tab:testres}. 
We can see that \mASR-\mdl{ComplEx} yields better results both in comparison with \mdl{KALE} and with the un-regularised model \mdl{ComplEx}.
When comparing \mdl{DistMult} with its \ASR-regularised extension, we note an improvement from 0.628 to 0.675 for MRR, and from 72.9\% to 75.2\% for Hits@10.
Similarly, when comparing \mdl{ComplEx} with its extension \mASR-\mdl{ComplEx}, we note an improvement from 0.641 to 0.698 for MRR, and from 71.9\% to 75.7\% for Hits@10.
Also note that \ASR, when used jointly with \mdl{ComplEx} and \mdl{DistMult}, yields larger relative improvements in comparison with \mdl{KALE} (a model inspired by \mdl{TransE}) and \mdl{TransE}.
Improvements are even more evident if we consider the FB122 results in Test-II, where test facts are directly related to the logic clauses.
We can see that, in terms of MRR, the proposed regularisation method improves the MRR from 0.874 to 0.948 for \mdl{DistMult}, and from 0.887 to 0.984 for \mdl{ComplEx}.
%
%

%
We ran the same experiments on WN18 (see the supplemental material for more exhaustive results) but did not notice any significant improvements.
One reason can be that the neural link prediction model has enough training data to \emph{learn} the WN18 rules by itself.
For this reason, we also evaluate \ASR in settings where the availability of training data is limited.
More specifically, at training time, we use a limited sample of the training triples (\ie{} 20\%, 30\%, \ldots) whose predicate appears in the head of a clause.
Results are available in Tab.~\ref{tab:limited}: we note that, in a limited data regime, \ASR yields some marginal improvements on WN18. 
In Tab.~\ref{tab:testres} we also report results for the closed-form solutions described in Sect.~\ref{ssec:closed} and derived in the Appendix, denoted by \mcASR-\mdl{ComplEx} and \mcASR-\mdl{DistMult}.
%
%
In both cases, the closed form solutions for the inner adversarial training loop yield similar results to their iterative counterparts, improving on all baselines.

\section{CONCLUSIONS}

In this paper, we introduced Adversarial Set Regularisation (\ASR), a general and scalable method for regularising neural link prediction models.
In the proposed method, an assumption is used for deriving an \emph{inconsistency loss} measuring the degree to which the model violates the assumption on an adversarially-trained set of input representations.
The training objective is defined as a zero-sum game, where an adversary finds the most offending set of input representations by maximising the inconsistency loss, and the model uses the inconsistency loss on such a set for regularising its training process.
Our results demonstrate that incorporating prior assumptions in the form of FOL clauses gives a steady improvement over neural link prediction models, especially when the availability of training data is limited.
Furthermore, the proposed method yields consistent improvements over the recently-proposed \mdl{KALE} method.
\vspace*{-5pt}
\subsubsection*{Acknowledgements}
\vspace*{-5pt}
We are immensely grateful to Jeff Mitchell, Johannes Welbl, and Jason Naradowsky for useful discussions.
This work has been supported by the Research Foundation - Flanders (FWO), Ghent University - iMinds, a Google PhD Fellowship in Natural Language Processing, an Allen Distinguished Investigator Award, and a Marie Curie Career Integration Award.

\vspace*{-5pt}
\bibliographystyle{abbrvnat}
\bibliography{references_short,references_rockt,references_tmp}
%


\newcommand{\lagr}{\ensuremath{\mathcal{L}}}

\newcommand{\re}[1]{\ensuremath{#1^{\text{R}}}}
\newcommand{\im}[1]{\ensuremath{#1^{\text{I}}}}

\global\long\def\x{\mathbf{\mathbf{x}}}
\global\long\def\y{\mathbf{\mathbf{y}}}
\global\long\def\h{\mathbf{\mathbf{h}}}
\global\long\def\hR{\mathbf{h^{\text{R}}}}
\global\long\def\hI{\mathbf{h^{\text{I}}}}

\global\long\def\r{\params_{r}}
\global\long\def\rR{\params_{r}^{\text{R}}}
\global\long\def\rI{\params_{r}^{\text{I}}}

\global\long\def\b{\params_{b}}
\global\long\def\bR{\params_{b}^{\text{R}}}
\global\long\def\bI{\params_{b}^{\text{I}}}

\global\long\def\z{\mathbf{z}}
\global\long\def\zR{\mathbf{z^{\text{R}}}}
\global\long\def\zI{\mathbf{z^{\text{I}}}}

\global\long\def\s{{\h_1}}
\global\long\def\sR{{\h_{1}^{\text{R}}}}
\global\long\def\sI{{\h_{1}^{\text{I}}}}
\global\long\def\o{{\h_2}}
\global\long\def\oR{{\h_2^{\text{R}}}}
\global\long\def\oI{{\h_2^{\text{I}}}}

\global\long\def\d{\boldsymbol{\delta}}
\global\long\def\dR{\boldsymbol{\delta^{\text{R}}}}
\global\long\def\dI{\boldsymbol{\delta^{\text{I}}}}
\global\long\def\1{\mathbf{1}}
\global\long\def\0{\mathbf{0}}
\global\long\def\muR{\mu^{\text{R}}}
\global\long\def\muI{\mu^{\text{I}}}
\global\long\def\lamR{\lambda^{\text{R}}}
\global\long\def\lamI{\lambda^{\text{I}}}

\global\long\def\hi{{h_i}}
\global\long\def\hRi{{h_i^{\text{R}}}}
\global\long\def\hIi{{h_i^{\text{I}}}}

\global\long\def\ri{{\params_{r, i}}}
\global\long\def\rRi{{\params_{r, i}^{\text{R}}}}
\global\long\def\rIi{{\params_{r, i}^{\text{I}}}}
\global\long\def\rRj{{\params_{r, j}^{\text{R}}}}
\global\long\def\rIj{{\params_{r, j}^{\text{I}}}}

\global\long\def\bi{{\params_{b, i}}}
\global\long\def\boi{{\params_{b, 1}_i}}
\global\long\def\bti{{\params_{b, 2}_i}}
\global\long\def\bRi{{\params_{b, i}^{\text{R}}}}
\global\long\def\bIi{{\params_{b, i}^{\text{I}}}}

\global\long\def\boRi{{\params_{b, 1}_i^{\text{R}}}}
\global\long\def\boIi{{\params_{b, 1}_i^{\text{I}}}}
\global\long\def\btRi{{\params_{b, 2}_i^{\text{R}}}}
\global\long\def\btIi{{\params_{b, 2}_i^{\text{I}}}}

\global\long\def\bRj{{\params_{b, j}^{\text{R}}}}
\global\long\def\bIj{{\params_{b, j}^{\text{I}}}}

\global\long\def\zi{{z_i}}
\global\long\def\zRi{{z_i^{\text{R}}}}
\global\long\def\zIi{{z_i^{\text{I}}}}
\global\long\def\deltai{{\delta_i}}
\global\long\def\deltaRi{{\delta_i^{\text{R}}}}
\global\long\def\deltaIi{{\delta_i^{\text{I}}}}
\global\long\def\zetaRi{{\zeta_i^{\text{R}}}}
\global\long\def\zetaIi{{\zeta_i^{\text{I}}}}

\global\long\def\si{{h_{1, i}}}
\global\long\def\sRi{{{h_{1, i}}^{\text{R}}}}
\global\long\def\sIi{{{h_{1, i}}^{\text{I}}}}
\global\long\def\oi{{h_{2, i}}}
\global\long\def\oRi{{{h_{2, i}}^{\text{R}}}}
\global\long\def\oIi{{{h_{2, i}}^{\text{I}}}}
\global\long\def\hoi{{h_{1, i}}}
\global\long\def\hti{{h_{2, i}}}

\global\long\def\mub{\boldsymbol{\mu}}
\global\long\def\deltab{\boldsymbol{\delta}}
\global\long\def\xib{\boldsymbol{\xi}}
\global\long\def\xibR{\boldsymbol{\xi}^{\text{R}}}
\global\long\def\xibI{\boldsymbol{\xi}^{\text{I}}}
\global\long\def\xiR{{\xi}^{\text{R}}}
\global\long\def\xiI{{\xi}^{\text{I}}}
\global\long\def\zetab{\boldsymbol{\zeta}}
\global\long\def\mubR{\boldsymbol{\mu}^{\text{R}}}
\global\long\def\deltabR{\boldsymbol{\delta}^{\text{R}}}
\global\long\def\zetabR{\boldsymbol{\zeta}^{\text{R}}}
\global\long\def\mubI{\boldsymbol{\mu}^{\text{I}}}
\global\long\def\deltabI{\boldsymbol{\delta}^{\text{I}}}
\global\long\def\zetabI{\boldsymbol{\zeta}^{\text{I}}}

\global\long\def\di{{\delta_i}}
\global\long\def\dRi{{\delta_i^{\text{R}}}}
\global\long\def\dIi{{\delta_i^{\text{I}}}}

\newpage

\onecolumn

\appendix

\begin{center}
\textbf{\textcolor{black}{\LARGE{}Supplementary Material}}
\par\end{center}{\LARGE \par}

\section{Closed Form Solutions for Maximum Violation Loss for Simple Implications}
Let $\fs_{r}(\h_1, \h_2)$ be a scoring function for a relation $r$ defined over pairs of entity vectors $\h_1$ and $\h_2$, such as the scoring function in \mdl{DistMult} or \mdl{ComplEx}.
In the following, we assume that all entity embeddings live on a subspace $\subspace \subseteq \Real^{k}$.
The subspace $\subspace$ can either correspond to the unit sphere -- \ie{} $\subspace \triangleq \{ \h \mid \norm{\h}_{2} = 1 \}$ -- or to the unit cube -- \ie{} $\subspace \triangleq \{ \h \mid \h \in \left[ 0, 1 \right]^{k} \}$.
Let us consider a mapping $\AdvSet : \vars \mapsto \Real^{k}$ from variables to $k$-dimensional embeddings, where $\h_i = \AdvSet(X_i)$, $\forall i$.
Given a clause expressing a simple implication in the form $b(X_1, X_2) \Implies r(X_1, X_2)$, we would like to maximise the inconsistency loss $\advLoss$ associated to the clause:
\begin{equation*}
 \begin{aligned} 
  & \advLossMax = \max\big(0, \lossMax \big)\\
  &\text{with }\quad \lossMax = \max_{\h_1, \h_2 \in \subspace} \big(\fs_{b}(\h_1, \h_2) - \fs_{r}(\h_1, \h_2) \big).
 \end{aligned}
\end{equation*}
The following sections show how to directly derive $\advLossMax$ for various models and entity embedding space restrictions.
One way to solve the maximisation problem is via the Karush–Kuhn–Tucker (KKT) conditions -- we refer to \citesup{Boyd:2004:CO:993483} for more details.

\subsection{\mdl{DistMult}}
In the following, we focus on the Bilinear-Diagonal model (\mdl{DistMult}), proposed by \citesup{yang15:embedding}, and provide the corresponding derivations for different choices of the entity embedding subspace $\subspace$.

We want to solve the following optimisation problem:
\begin{equation*}
\begin{aligned}
\lossMax &= \max_{\h_1, \h_2 \in \subspace} \big(\fs_{b}(\h_1, \h_2) - \fs_{r}(\h_1, \h_2) \big)\\
&= \max_{\h_1, \h_2 \in \subspace} \tdot{\b}{\h_1}{\h_2} - \tdot{\r}{\h_1}{\h_2}\\
&= \max_{\h_1, \h_2 \in \subspace} \tdot{\deltab}{\h_1}{\h_2},
\end{aligned}
\end{equation*}
where $\deltab \triangleq \b - \r$.

\subsubsection{Unit Sphere}
Assume that the subspace $\subspace$ corresponds to a the unit sphere, \ie{} $\forall x \in \ents : \norm{\h_x}_{2}^{2} = 1$.
The Lagrangian is:
\begin{equation*}
\lagr = - \tdot{\deltab}{\h_1}{\h_2} + \lambda_1(\norm{\h_1}_2^2-1) + \lambda_2(\norm{\h_2}_2^2-1).
\end{equation*}
Imposing stationarity: $\nabla_{\h_1} \lagr = 0$ and $\nabla_{\h_2} \lagr = 0$ gives:
\begin{equation*}
\begin{aligned}
-\deltab\odot\h_2 + 2\lambda_1\h_1 = 0\\
-\deltab\odot\h_1 + 2\lambda_2\h_2 = 0
\end{aligned}
\end{equation*}
in which $\odot$ denotes the component-wise multiplication.
For $\lambda_1\not=0$, a simple substitution leads to:
\begin{equation*}
-\deltab^2\odot\h_2 + 4\lambda_1\lambda_2\h_2 = 0,
\end{equation*}
with the notation $\deltab \odot \deltab = \deltab^2$.
As a result, $4\lambda_1\lambda_2=\delta_i^2$ for components $i$ with $\hoi\not=0$.
Given the symmetry of the equations, the same requirements $4\lambda_1 \lambda_2 = \delta_i^2$ hold for components $\hti\not=0$.
We search for $\h_1$ and $\h_2$, such that  $\delta_i^2$ is constant for their non-zero components.
Construct $\h_1$ and $\h_2$ such that only their component $j$ is non-zero:
$\forall i\not=j: \hoi=\hti=0$, whereas $h_{1, j} = \pm 1$, $h_{2, j} = \pm 1$ (given the unit sphere constraint).
The contribution of component $j$ to $\left\langle \deltab, \s,\o\right\rangle$ depends on $\hoi\hti$ which can take values $\pm 1$.
As a result:
\begin{equation*}
\begin{aligned}
 \lossMax &= \max_j \vert \delta_j \vert.
\end{aligned}
\end{equation*}
In the case where several components $\delta_j$ have the same value, then all $\hoi$ and $\hti$ need to be zero for $i\not=j$, but due to the normalisation constraint, the highest value of $\sum_j h_{1_j}h_{2_j}$ is found for a single index $j$ if both components take the value $\pm 1$.

Finally, since $\lossMax$ is always non-negative, we find:
\begin{equation*}
\boxed{\advLossMax = \max_j \vert \params_{b, j} - \params_{r, j} \vert}
\end{equation*}
\subsubsection{Unit cube}
For the entity embeddings to be constrained in the unit cube, their subspace is set to $\subspace = [0, 1]^{k} \subset \Real^{k}$.
This corresponds to reducing the entity embeddings to approximately Boolean embeddings(see \citesup{DBLP:conf/emnlp/DemeesterRR16}).

The Lagrangian becomes:
\begin{equation*}
\begin{aligned}
\lagr =  -\tdot{\deltab}{\h_1}{\h_2} + \sum_i\left[\mub_1\odot (\h_1 - \1) + \mub_2\odot (\h_2 - \1)\right]_i
\end{aligned}
\end{equation*}
with $\forall i : \hoi - 1 \leq 0, \hti - 1 \leq 0$ (primal feasibility), $\forall i: \mu_{1, i} \geq 0, \mu_{2, i} \geq 0$ (dual feasibility), $\forall i : \mu_{1, i} (\hoi - 1) = 0,\; \mu_{2, i}(\hti - 1) = 0$ (complementary slackness).
In fact, we should add KKT multipliers for the conditions $-\hoi\leq 0, -\hti\leq 0$ as well. These don't change the results, if we ensure the unit cube restrictions are satisfied.

Imposing stationarity, or $\nabla_{\h_1} \lagr = 0$ and $\nabla_{\h_2} \lagr = 0$, we get:
\begin{equation*}
 \begin{aligned}
  -\deltab\odot\h_2 + \mub_1 = 0\\
  -\deltab\odot\h_1 + \mub_2 = 0
 \end{aligned}
\end{equation*}

This can be solved component-wise, or for any component $i$:
\begin{equation*}
 \begin{aligned}
  {\mu_1}_i = \delta_i \hti \\
  {\mu_2}_i = \delta_i \hoi
 \end{aligned}
\end{equation*}
Dual feasibility dictates that ${\mu_1}_i\geq 0$ and ${\mu_2}_i\geq 0$, such that $\hti=0$ and $\hoi=0$ for any components where $\delta_i<0$.
For the other components, we have $\delta_i\geq 0$, such that while satisfying the unit cube constraints the highest value of the objective becomes
$\lossMax_i = \delta_i$ for $\hoi=\hti=1$.

Finally, we find:
\begin{equation*}
\boxed{
 \advLossMax = \sum_j \max(0, \params_{b, j} - \params_{r, j}),
}
\end{equation*}
which is exactly the same expression as the lifted loss for simple implications with unit cube entity embeddings introduced by \citesup{DBLP:conf/emnlp/DemeesterRR16} for \mdl{Model F}  (which can be seen as a special case of \mdl{DistMult} where the subject embeddings are replaced by entity pair embeddings, and all object embedding components are set to 1).

\subsection{\mdl{ComplEx}}
In \mdl{ComplEx}, proposed by \citesup{DBLP:conf/icml/TrouillonWRGB16}, the scoring function $\fs$ is defined as follows:
\begin{equation*}
 \begin{aligned}
\fs_r^{\params}(\h_1, \h_2) \triangleq \; & \tdot{\r}{\h_1}{\conjt{\h_2}}^{\text{R}} 
 \end{aligned}
\end{equation*}
\noindent where $\x^{\text{R}} \triangleq \ReP{\x}$ and $\x^{\text{I}} \triangleq \ImP{\x}$ denote the real and imaginary part of $\x$, respectively.
In the following, we analyse the cases where entity embeddings live on the unit sphere (\ie{} $\forall x \in \ents : \norm{\h_{x}}_{2}^{2} = 1$) and in the unit cube (\ie $\forall x \in \ents : \h_{x}^\text{R} \in [0, 1]^{k}, \; \h_{x}^\text{I} \in [0, 1]^{k}$).

We want to solve the following maximisation problem:
\begin{equation*}
 \begin{aligned}
  \loss &= \max_{\h_1, \h_2 \in \subspace} \big(\fs_{b}(\h_1, \h_2) - \fs_{r}(\h_1, \h_2) \big)\\
 &= \max_{\h_1, \h_2 \in \subspace} \tdot{\deltab}{\s}{\overline\o}^{\text{R}},
 \end{aligned}
\end{equation*}
with the complex vector $\deltab= \b - \r$.

To keep notations simple, we will continue to work with complex vectors as much as possible.
For the Lagrangian $\lagr$, the pair of stationarity equations with respect to the real and imaginary part of a variable (say $\s$), can be taken together as follows:
\[
\nabla_{\sR}\lagr + j\nabla_{\sI}\lagr = 0
\]
and we introduce the notation $\nabla_{\s}\lagr \triangleq \nabla_{\sR}\lagr + j\nabla_{\sI}\lagr$, with $j$ the imaginary unit.
For the remainder, we need the following:
\begin{equation*}
\begin{aligned}
\nabla_{\s} \tdot{\deltab}{\s}{\overline\o}^{\text{R}}
&= \frac{1}{2} \nabla_{\s} \left( \tdot{\deltab}{\s}{\overline\o} + \tdot{\overline\deltab}{\overline\s}{\o}\right)\\
&= \frac{1}{2} \left( \deltab\odot\overline\o + j\left(j\deltab\odot\overline\o\right)
+ \overline\deltab\odot\o + j\left(-j\overline\deltab\odot\o\right)
\right)\\
&=\overline\deltab\odot\o\\
\nabla_{\o} \tdot{\deltab}{\s}{\overline\o}^{\text{R}}
&= \deltab\odot\s
\end{aligned}
\end{equation*}
\subsubsection{Unit sphere}
We now restrict the subject and object embeddings to live on the unit sphere, \ie{} $\forall x \in \ents : \norm{\h_{x}}_{2}^{2} = 1$.
Given that the adversarial entity embeddings need to live on the unit sphere, the Lagrangian can be defined as follows:
\begin{equation*}
\begin{aligned}
\lagr(\s,\o,\lambda_1,\lambda_2) =
& - \tdot{\deltab}{\s}{\overline\o}^\text{R}\\
& + \lambda_1 \left( \norm{\s}_2^2 - 1 \right)\\
& + \lambda_2 \left( \norm{\o}_2^2 - 1 \right)
\end{aligned}
\end{equation*}
with real-valued Lagrange multipliers $\lambda_1$ and $\lambda_2$, and in which $\norm{\s}_2^2 \triangleq \norm{\sR}_2^2 + \norm{\sI}_2^2$ is the L2 norm of the complex vector $\s$.
With the expressions for $\nabla_{\s} \tdot{\deltab}{\s}{\overline\o}^{\text{R}}$ and $\nabla_{\o} \tdot{\deltab}{\s}{\overline\o}^{\text{R}}$ above, the stationarity conditions can be written out as follows:
\begin{equation*}
\begin{aligned}
-\overline{\deltab} \odot \o + 2 \lambda_1\s = 0 \\
-\deltab\odot \s + 2\lambda_2\o = 0
\end{aligned}
\end{equation*}
Substituting into each other, we find:
\begin{equation*}
\begin{aligned}
4\lambda_1\lambda_2 \s
&= \overline\deltab\odot\deltab\odot\s,\qquad(\lambda_2\not=0)\\
4\lambda_1\lambda_2 \o
&= \deltab\odot\overline\deltab\odot\o,\qquad(\lambda_1\not=0)
\end{aligned}
\end{equation*}
As a result, we require $4\lambda_1\lambda_2=\vert\delta_i\vert^2$ for components $i$ with $\hoi\not=0$ or $\hti\not=0$.
As in the case with DistMult, take $\hoi=\hti = 0$ for each component $i\not=j$, such that for component $j$, we need $\vert h_{1, j} \vert = \vert h_{2, j} \vert = 1$, such that $\vert h_{1, j} h_{2, j} \vert = 1$.
In order to maximise the contribution of that component to the loss, we choose the argument of the complex number $h_{1, j} \overline{h_{2, j}}$ such that $\delta_j h_{1, j} \overline{h_{2, j}}$ falls on the positive real axis.
As a result:
\begin{equation*}
\boxed{
 \advLossMax = \max_j \vert b_j - r_j \vert = \max_j \sqrt{(\bRj-\rRj)^2 + (\bIj-\rIj)^2 }
}
\end{equation*}
\subsubsection{Unit cube}
This case can be solved with the KKT conditions again, but instead we provide a shorter, less formal, derivation.
It is clear that we can maximise the objective by maximising each component independently.
For component $i$ we need to optimise the following:
\begin{equation*}
 \begin{aligned}
  \deltaRi \sRi \oRi + \deltaRi \sIi \oIi + \deltaIi \sRi \oIi - \deltaIi \sIi \oRi
 \end{aligned}
\end{equation*}
Regrouping gives:
\begin{equation*}
 \begin{aligned}
  \alpha \deltaRi + \beta \deltaIi,
 \end{aligned}
\end{equation*}
with $\alpha = \sRi \oRi + \sIi \oIi$ and $\beta = \sRi \oIi - \sIi \oRi$. We know $0 \leq \alpha \leq 2$, $-1 \leq \beta \leq 1$ and $\alpha + |\beta| \leq 2$.

This allows maximising the objective as follows:
\begin{equation*}
\boxed{
 \advLossMax = \sum_i \max(\deltaRi,0) + \max(\deltaRi,\vert\deltaIi\vert)
}
\end{equation*}
\section{Simple Implications with Swapped Arguments} \label{app:impl_inv}
Given a clause expressing a simple implication with swapped arguments, in the form $b(X_1, X_2) \Implies r(X_2, X_1)$, we would like to maximise the inconsistency loss $\advLoss$ associated to the clause, \ie{}:
\begin{equation*}
 \begin{aligned}
  & \advLossMax = \max\big(0, \lossMax \big)\\
  &\text{with }\quad \lossMax = \max_{\h_1, \h_2 \in \subspace} \big(\fs_{b}(\h_1, \h_2) - \fs_{r}(\h_2, \h_1) \big).
 \end{aligned}
\end{equation*}
\subsection{\mdl{DistMult}}
Due to symmetry, the same close form expressions as for the simple implications hold.
\subsection{\mdl{ComplEx}}
We want to solve the following maximisation problem:
\begin{equation*}
 \begin{aligned}
  \lossMax &= \max_{\h_1, \h_2 \in \subspace} \big(\fs_{b}(\h_1, \h_2) - \fs_{r}(\h_2, \h_1) \big)\\
  &= \max_{\h_1, \h_2 \in \subspace} \tdot{\b}{\s}{\overline\o}^{\text{R}} - \tdot{\r}{\o}{\overline\s}^{\text{R}}\\
  &= \max_{\h_1, \h_2 \in \subspace}
  \tdot{\b}{\s}{\overline\o}^{\text{R}} - \tdot{\overline\r}{\overline\o}{\s}^{\text{R}}\\
  &= \max_{\h_1, \h_2 \in \subspace}
  \tdot{\b-\overline\r}{\s}{\overline\o}^{\text{R}}\\
  &= \max_{\h_1, \h_2 \in \subspace}
  \tdot{\zetab}{\s}{\overline\o}^{\text{R}}
 \end{aligned}
\end{equation*}
This has the same form as the simple implications case, but with $\r$ replaced by $\overline\r$, or, more specifically, $\im{\r}$ by $-\im{\r}$.
\subsubsection{Unit sphere}

Under unit sphere constraints, $\advLossMax$ has the following value:
\begin{equation} \label{eq:complexsphere}
\boxed{
\advLossMax = \max_i \sqrt{(\bRi-\rRi)^2 + (\bIi+\rIi)^2 }
}
\end{equation}

\subsubsection{Unit cube}
With $\re{\zeta}_i=\bRi-\rRi$ and $\im{\zeta}_i=\bIi+\rIi$:
\begin{equation*}
\boxed{
\advLossMax = \sum_i \max(\zetaRi,0) + \max(\zetaRi,\vert\zetaIi\vert)
}
\end{equation*}

\section{Symmetry}
Given a clause expressing a simple implication with swapped arguments, in the form $r(X_1, X_2) \Implies r(X_2, X_1)$, we would like to maximise the inconsistency loss $\advLoss$ associated to the clause, \ie{}:
\begin{equation*}
 \begin{aligned}
  & \advLossMax = \max\big(0, \lossMax \big)\\
  &\text{with }\quad \lossMax = \max_{\h_1, \h_2 \in \subspace} \big(\fs_{r}(\h_1, \h_2) - \fs_{r}(\h_2, \h_1) \big).
 \end{aligned}
\end{equation*}
Note that this is a special case of Appendix \ref{app:impl_inv} where $r = b$.

\subsection{\mdl{DistMult}}
Since \mdl{DistMult} is symmetric, the gradient for symmetry clauses is zero, \ie{}, all relations already satisfy symmetry.

\subsection{\mdl{ComplEx}}
We want to solve the following maximisation problem:
\begin{equation*}
 \begin{aligned}
  \lossMax &= \max_{\h_1, \h_2 \in \subspace} \big(\fs_{r}(\h_1, \h_2) - \fs_{r}(\h_2, \h_1) \big)\\
  &= \max_{\h_1, \h_2 \in \subspace} \tdot{\r-\overline\r}{\s}{\overline\o}^{\text{R}}
 \end{aligned}
\end{equation*}
\subsubsection{Unit sphere}
From Eq. (\ref{eq:complexsphere}) with $\rRi = \bRi$ and $\rIi = \bIi$ we get:
\begin{equation*}
\boxed{
\lossMax = \max_i 2|\rIi|
}
\end{equation*}

\subsubsection{Unit cube}
Similarly, with $\rRi = \bRi$ and $\rIi = \bIi$ we get:
\begin{equation*}
\boxed{\lossMax = 2\sum_i|\rIi|
}
\end{equation*}

\section{Link Prediction Results}

\begin{table*}[h!]
\centering
\caption{Link prediction results on the Test-I, Test-II and and Test-ALL on FB122, filtered setting.}
\begin{adjustbox}{width=1\textwidth}
\begin{tabular}{C{1cm}lC{1cm}C{1cm}C{1cm}C{1cm}C{.5cm}C{1cm}C{1cm}C{1cm}C{1cm}C{.5cm}C{1cm}C{1cm}C{1cm}C{1cm}}
\toprule
& & \multicolumn{4}{c}{Test-I} & & \multicolumn{4}{c}{Test-II} & & \multicolumn{4}{c}{Test-ALL} \\
\cline{3-6}\cline{8-11}\cline{13-16}
& & \multicolumn{3}{c}{Hits@N (\%)} & \multirow{2}{*}{MRR} & & \multicolumn{3}{c}{Hits@N (\%)} & \multirow{2}{*}{MRR} & & \multicolumn{3}{c}{Hits@N (\%)} & \multirow{2}{*}{MRR} \\
\cline{3-5}\cline{8-10}\cline{13-15}
& & 3 & 5 & 10 & & \ & 3 & 5 & 10 & \ & & 3 & 5 & 10 & \ \\
\cline{2-16}
\multirow{9}{*}{\rotatebox[origin=c]{90}{\parbox[c]{1cm}{\centering {\bf FB122}}}}
& \mdl{TransE}~\citep{DBLP:conf/nips/BordesUGWY13} & 36.0 & 41.5 & 48.1 & 0.296 & & 77.5 & 82.8 & 88.4 & 0.630 & & 58.9 & 64.2 & 70.2 & 0.480 \\
& \mdl{KALE-Pre}~\citep{DBLP:conf/emnlp/GuoWWWG16} & 35.8 & 41.9 & 49.8 & 0.291 & & 82.9 & 86.1 & 89.9 & 0.713 & & 61.7 & 66.2 & 71.8 & 0.523 \\
& \mdl{KALE-Joint}~\citep{DBLP:conf/emnlp/GuoWWWG16} & {\bf 38.4} & {\bf 44.7} & {\bf 52.2} & 0.325 & & 79.7 & 84.1 & 89.6 & 0.684 & & 61.2 & 66.4 & 72.8 & 0.523 \\
\cline{2-16}
%

& \mdl{DistMult}~\citep{yang15:embedding} & 36.0 & 40.3 & 45.3 & 0.313 & & 92.3 & 93.8 & 94.7 & 0.874 & & 67.4 & 70.2 & 72.9 & 0.628 \\
& \emph{ASR}-\mdl{DistMult} & 36.3 & 40.3 & 44.9 & 0.330 & & 98.0 & 99.0 & 99.2 & 0.948 & & 70.7 & 73.1 & 75.2 & 0.675 \\
& \emph{cASR}-\mdl{DistMult} & 37.0 & 40.4 & 45.1 & 0.337 & & 96.7 & 98.6 & 99.3 & 0.933 & & 70.1 & 72.7 & 75.1 & 0.669 \\
\cline{2-16}
%

& \mdl{ComplEx}~\citep{DBLP:conf/icml/TrouillonWRGB16} & 37.0 & 41.3 & 46.2 & 0.329 & & 91.4 & 91.9 & 92.4 & 0.887 & & 67.3 & 69.5 & 71.9 & 0.641 \\
& \emph{ASR}-\mdl{ComplEx} & 37.3 & 41.0 & 45.9 & 0.338 & & {\bf 99.2} & {\bf 99.3} & {\bf 99.4} & {\bf 0.984} & & {\bf 71.7} & {\bf 73.6} & {\bf 75.7} & {\bf 0.698}
\\
& \emph{cASR}-\mdl{ComplEx} & 37.9 & 41.7 & 46.2 & {\bf 0.339} & & 97.7 & {\bf 99.3} & {\bf 99.4} & 0.954 & & 71.1 & {\bf 73.6} & 75.6 & 0.680 \\
\midrule
\midrule
\multirow{7}{*}{\rotatebox[origin=c]{90}{
 \parbox[c]{1cm}{\centering {\bf WN18}}}
}
& \mdl{TransE}~\citep{DBLP:conf/nips/BordesUGWY13} & 57.4 & 72.3 & 80.1 & 0.306 & & 87.5 & 95.6 & 98.7 & 0.511 & & 79.1 & 89.1 & 93.6 & 0.453 \\
& \mdl{KALE-Pre}~\citep{DBLP:conf/emnlp/GuoWWWG16} & 60.6 & 74.5 & 81.1 & 0.322 & & 96.4 & 98.6 & 99.6 & 0.612 & & 86.4 & 91.9 & 94.4 & 0.532 \\
& \mdl{KALE-Joint}~\citep{DBLP:conf/emnlp/GuoWWWG16} & 65.5 & 76.3 & 82.1 & 0.338 & & 93.3 & 95.4 & 97.2 & 0.787 & & 85.5 & 90.1 & 93.0 & 0.662 \\
\cline{2-16}
& \mdl{DistMult}~\citep{yang15:embedding} & 80.6 & 81.6 & 82.6 & 0.796 & & 96.7 & 98.4 & 99.5 & 0.872 & & 92.3 & 93.7 & 94.9 & 0.850 \\
& \emph{ASR}-\mdl{DistMult} & {\bf 81.4} & {\bf 82.0} & {\bf 82.9} &  0.801 & & 96.7 & 98.4 & 99.5 &  0.869 & & 92.4 & 93.8 & 94.9 & 0.851 \\
\cline{2-16}
& \mdl{ComplEx}~\citep{DBLP:conf/icml/TrouillonWRGB16} & 81.0 & 81.8 & 82.5 & {\bf 0.803} & & {\bf 99.9} & {\bf 100} & {\bf 100} & {\bf 0.996} & & {\bf 94.7} & {\bf 95.0} & {\bf 95.1} & {\bf 0.942} \\
& \emph{ASR}-\mdl{ComplEx} & 81.0 & 81.8 & 82.5 & {\bf 0.803} & & {\bf 99.9} & {\bf 100} & {\bf 100} & {\bf 0.996} & & {\bf 94.7} & {\bf 95.0} & {\bf 95.1} & {\bf 0.942} \\
\bottomrule
\end{tabular}
\end{adjustbox}
\end{table*}

\bibliographystylesup{abbrvnat}
\bibliographysup{refappendix}

\end{document}